\documentclass{article} 
\usepackage{iclr2025_conference,times}

\usepackage{wrapfig}
\usepackage{lipsum} 
\usepackage{wrapfig}

\usepackage{amsmath,amsfonts,bm}
\usepackage{multicol}

\def\eqref#1{equation~\ref{#1}}

\def\1{\bm{1}}

\DeclareMathAlphabet{\mathsfit}{\encodingdefault}{\sfdefault}{m}{sl}
\SetMathAlphabet{\mathsfit}{bold}{\encodingdefault}{\sfdefault}{bx}{n}

\usepackage{booktabs}      
\usepackage{multirow}      
\usepackage{amssymb}       
\usepackage{graphicx, subfigure}     

\usepackage{colortbl}
\usepackage{xcolor}

\usepackage{hyperref}
\usepackage{url}
\usepackage{multicol}
\usepackage{aliascnt}
\usepackage{adjustbox}
\usepackage{siunitx} 
\usepackage{booktabs}
\usepackage{multirow} 
\usepackage{enumitem}
\usepackage{booktabs} 
\usepackage{amssymb}  
\usepackage{amsmath}  
\usepackage{graphicx} 
\usepackage{longtable} 
\usepackage{graphicx}
\usepackage{subcaption}
\usepackage{calc}
\usepackage[most]{tcolorbox}
\usepackage{listings}
\usepackage{latexsym}
\usepackage{dsfont}

\usepackage[ruled]{algorithm2e}

\definecolor{uclablue}{rgb}{0.15, 0.45, 0.68}
\hypersetup{
    breaklinks,
    citecolor=uclablue,
    colorlinks=true,
}

\usepackage{tabularx}
\robustify\bfseries
\usepackage{subcaption}
\usepackage{cleveref}

\tcbuselibrary{listings,skins,breakable}

\newtcolorbox{prompt}[1][]{enhanced,
  breakable,
  colback=white,              
  colframe=black,             
  coltitle=black,             
  colbacktitle=gray!20,       
  fonttitle=\bfseries,
  title=Prompt,
  #1}

\definecolor{linkColor}{rgb}{0.2,0.4,0.6}
\definecolor{myblue}{HTML}{0379AC}
\definecolor{myred}{HTML}{A50E50}
\definecolor{myorange}{RGB}{238, 133, 74}
\definecolor{latentcolor}{named}{cyan}
\definecolor{normalcolor}{RGB}{0, 0, 0}
\usepackage{marvosym}

\usepackage{xspace}

\title{LaSeR: Reinforcement Learning with Last-Token Self-Rewarding}

\author{
Wenkai Yang$^{1,}$\thanks{\ Work done during an internship at Tencent.} \ , Weijie Liu$^{2}$, Ruobing Xie$^{2}$, Yiju Guo$^1$, \\ Lulu Wu$^2$, Saiyong Yang$^2$, Yankai Lin$^{1,}$\thanks{\ Corresponding author.}\\ 
\textbf{$^1$Gaoling School of Artificial Intelligence, Renmin University of China} \\ \textbf{$^2$LLM Department, Tencent}\\
\Letter~\{wenkaiyang,yankailin\}@ruc.edu.cn\\
}

\begin{document}
\maketitle
\let\oldthefootnote\thefootnote

\let\thefootnote\oldthefootnote


\begin{abstract}
Reinforcement Learning with Verifiable Rewards (RLVR) 
has recently emerged as a core paradigm for enhancing the reasoning capabilities of Large Language Models (LLMs). 
To address the lack of verification signals at test time, prior studies incorporate the training of model's self-verification capability into the standard RLVR process, thereby unifying reasoning and verification capabilities within a single LLM. However, previous practice requires the LLM to sequentially generate solutions and self-verifications using two separate prompt templates, which significantly reduces efficiency. 
In this work, we theoretically reveal that the closed-form solution to the RL objective of self-verification can be reduced to a remarkably simple form: \textbf{the true reasoning reward of a solution is equal to its \textit{last-token self-rewarding score}}, which is computed as the difference between the policy model's next-token log-probability assigned to any pre-specified token at the solution's last token and a pre-calculated constant, scaled by the KL coefficient. Based on this insight, we propose \textbf{LaSeR} (Reinforcement Learning with \underline{La}st-Token \underline{Se}lf-\underline{R}ewarding), an algorithm that simply augments the original RLVR loss with a MSE loss that aligns the last-token self-rewarding scores with verifier-based reasoning rewards, jointly optimizing the reasoning and self-rewarding capabilities of LLMs. 
The optimized self-rewarding scores can be utilized in both training and testing to enhance model performance. 
Notably, our algorithm derives these scores from the predicted next-token probability distribution of the last token immediately after generation, incurring only the minimal extra cost of one additional token inference. 
Experiments show that our method not only improves the model's reasoning performance but also equips it with remarkable self-rewarding capability, thereby boosting its inference-time scaling performance. Code and models are available at \url{https://github.com/RUCBM/LaSeR}.
\end{abstract}

\begin{figure*}[t] 
    \centering
    \includegraphics[width=0.98\textwidth]{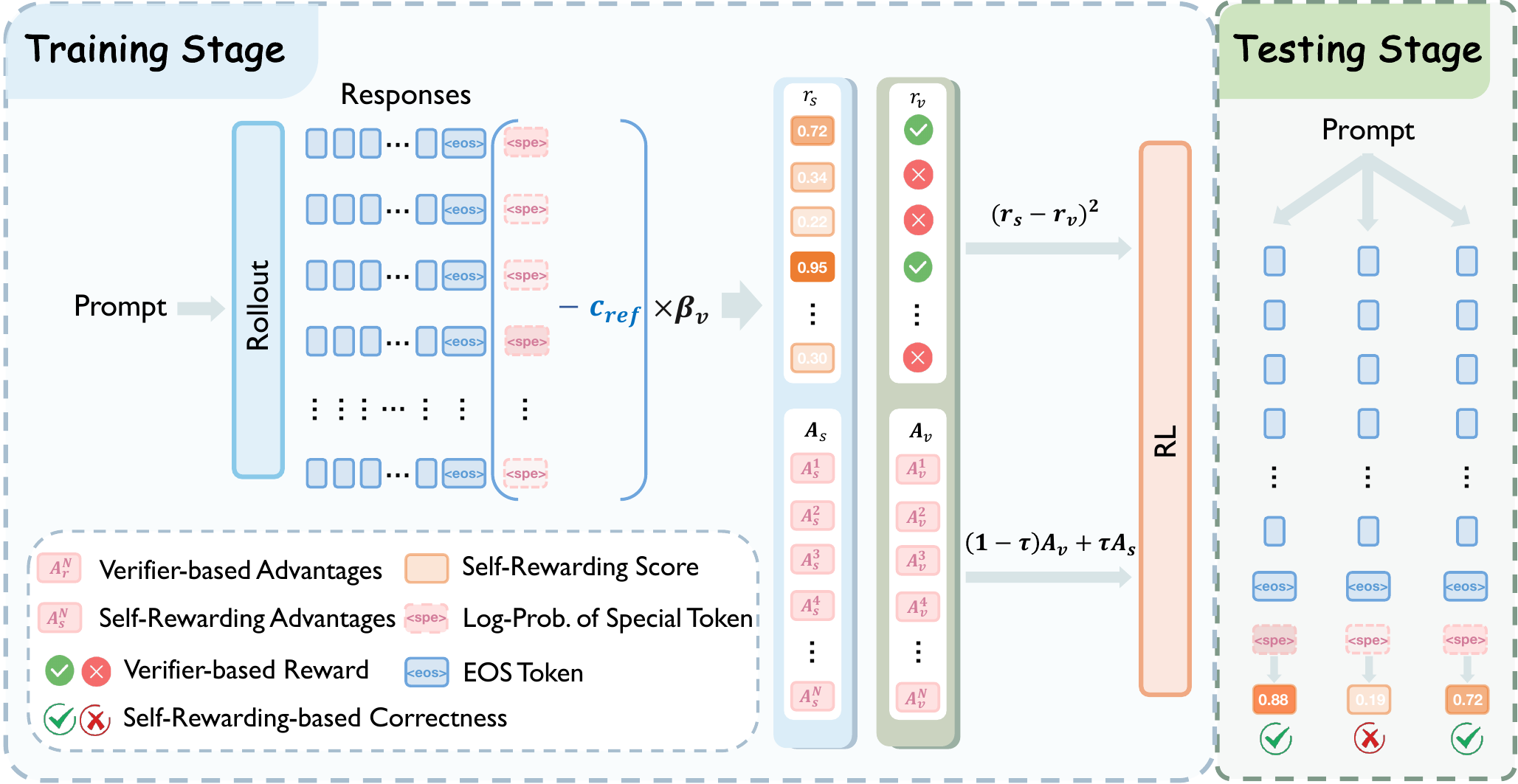}
    \caption{The full illustration of our method \textbf{LaSeR}. During training, our approach augments the standard RLVR process with an additional MSE loss between the verifier-based rewards ($r_{v}$) and the last-token self-rewarding scores ($r_{s}$), where $r_{s}$ is the difference between the policy model’s next-token log-probabilities of a pre-specified special token at the final response token and a pre-calculated constant $c_{ref}$, scaled by the KL coefficient $\beta_{v}$. The optimized self-rewarding scores can serve as auxiliary reward signals in both training and testing stages to enhance model performance.
    }
    \label{fig: pipeline}
\end{figure*}

\section{Introduction}
In the past few years, Large Language Models (LLMs)~\citep{gpt4,llama3.1,qwen2.5,deepseek-v3} have advanced significantly, excelling in various domains~\citep{alpaca_eval,mmlu-pro}. However, they still face limitations in complex reasoning tasks~\citep{aime2024,aime2025,gpqa,livecodebench}. Recently, Reinforcement Learning with Verifiable Rewards (RLVR) has shown great promise in enhancing the complex reasoning abilities of LLMs, as demonstrated by OpenAI o1~\citep{o1} and DeepSeek-R1~\citep{r1}. By rewarding reasoning paths based on the consistency between final outcomes and ground-truth answers through a deterministic verifier, RLVR incentivizes LLMs to produce more deliberate reasoning chains while effectively mitigating the risk of reward hacking~\citep{reward_hacking}.

Despite its effectiveness, a limitation of standard RLVR is its inability to continue providing verification signals for model outputs in scenarios where ground truth answers are unavailable, such as during test-time inference~\citep{TTRL}. To address this, existing works either train an external verifier~\citep{prm800k,scaling-test-time-compute,generative_verifiers,llm-critic-catch-math-bugs,deepcritic} for evaluating candidate solutions or jointly optimize the self-verification and reasoning capabilities of the same policy model during RLVR~\citep{put-value-back-in-rl,trust-but-verify,rl-tango,pag}. However, we argue that \textbf{these methods have a major issue of inefficiency}: the external verifier requires additional training on a separate LLM during or after reinforcement learning (RL); while joint optimization involves generating both solutions and self-verifications sequentially under two separate prompt templates, which doubles the per-sample inference cost and reduces generation efficiency.~\looseness=-1

In this work, we propose \textbf{LaSeR} (Reinforcement Learning with \underline{La}st-Token \underline{Se}lf-\underline{R}ewarding), a lightweight and highly effective algorithm that achieves this goal, jointly optimizing reasoning and self-verification capabilities at nearly zero additional cost. Our core insight is that a model's assessment in its own solution can be captured in the last token's predicted probability distribution. 
We first show theoretically that the RL objective of self-verification has a closed-form solution, where the true reasoning reward from the verifier is equal to the next-token log-probability ratio between the policy and reference models for a pre-specified special token (an unused token like ``\texttt{<|vision\_start|>}'' that serves as the pre-defined ground truth for verifications on correct candidate solutions) at the last response token, scaled by the KL coefficient. We refer to this scaled log-probability ratio  as the \textit{last-token self-rewarding score}. 
Furthermore, we observe that for a randomly chosen special token, its predicted log-probability under the reference model is practically a constant, small value across all problems and solutions (see Figure~\ref{fig: small value of log ref}). This enables us to simplify the self-rewarding score into a remarkably simple form that depends only on the policy model's outputs and a pre-calculated constant, making it exceptionally efficient to compute.

Building on above analysis, we replace the explicit RL optimization for self-verification with a simple Mean Squared Error (MSE) loss. As illustrated in Figure~\ref{fig: pipeline}, we train the model to align its last-token self-rewarding score with the true reasoning reward from the verifier. In specific, after the policy model generates the solution for each problem, we calculate the last-token self-rewarding score based on its last token's next-token log-probability for the pre-specified special token, and construct the corresponding MSE loss. This MSE objective is added directly to the standard RLVR loss, allowing for seamless joint optimization for both the reasoning and self-rewarding capabilities of the policy model. At both training and testing time, our method generates each candidate solution and computes the self-rewarding score in a single forward pass, incurring the cost of at most one additional token inference with no extra generation required. This is significantly more efficient than prior approaches, which require a separate inference step. The optimized self-rewarding scores can not only complement the original reasoning rewards during RLVR to further enhance training performance, but also be used at test time to rank and weight solutions for more accurate answer aggregation.~\looseness=-1

We conduct experiments on both LLaMA~\citep{llama3.2} and Qwen~\citep{qwen2.5} architectures, including pre-trained, mid-trained and reinforced variants, to demonstrate the effectiveness of our method in broader math reasoning tasks. Experimental results show that our methods not only effectively improve the reasoning performance of the policy model, but also allows its self-rewarding accuracy to reach a high level, thereby equipping the model with better confidence calibration of its own outputs and improving its inference-time scaling performance.

\section{Related Work}
\textbf{RLVR for LLM Reasoning} Reinforcement Learning with Verifiable Rewards (RLVR), which sorely calculates binary rewards based on the final answers, has been shown to be highly effective in enhancing the reasoning capabilities of LLMs~\citep{o1, r1, k1.5}. Current studies can be categorized into several directions, including but not limited to (1) designing more efficient and effective RLVR algorithms~\citep{ppo,deepseekmath,dapo,vapo,dr-grpo,gspo}, (2) extending RLVR to general reasoning domain~\citep{general-reasoner,verifree,rlpr,star-r1} and agent scenarios~\citep{ragen,k2, arpo}, (3) collecting diverse verifiable datasets~\citep{orz,deepmath,code-r1,general-reasoner,megascience}, and (4) analyzing the mechanisms of RLVR~\citep{rl-finetunes-small-net,passk,rlvr-correct-cot,rl-transferability}.~\looseness=-1

\textbf{External Verifiers for LLM Reasoning} Training external verifiers to identify the correctness of the LLM-generated solutions is an effective way to enhance the reasoning performance of LLMs in the inference time. External verifiers usually fall into two categories: (1) \textbf{Scalar Reward Models}: Outcome-supervised Reward Models~(ORMs)~\citep{gsm8k,qwen2.5-math} and Process-supervised Reward Models~(PRMs)~\citep{prm800k,math-shepherd,skywork-o1,implicitprm} are two representative approaches. ORMs provide supervision by evaluating the final answer, while PRMs offer more fine-grained feedback by assessing the intermediate reasoning steps. 
(2) \textbf{Generative Verifiers}: Recent studies have explored the potential of training LLMs to perform natural language critiques of reasoning solutions generated by the LLM generators, and then to judge their final outcomes~\citep{generative_verifiers, llm-critic-catch-math-bugs, deepcritic, genprm}. This paradigm has demonstrated stronger verification performance than scalar reward models, as it enables the LLM verifier to conduct deliberate chain-of-thought reasoning before arriving at the final judgment.~\looseness=-1

\textbf{Self-Verification for LLM Reasoning} 
Several recent studies~\citep{put-value-back-in-rl,trust-but-verify,rl-tango,pag,urpo} aim to unify the roles of generator and verifier by equipping a single policy model with self-verification capability. The trained self-verification capability can be used in both the RL training and inference-time scaling stages to enhance the model performance. However, these approaches require generating solutions and self-verifications sequentially during training and inference. In contrast, our method derives the self-rewarding signal directly from the next-token probability distribution of the final token of the generated sequence, achieving a more efficient and effective unification of generation and self-verification.

\section{Methodology}
\subsection{Preliminaries}
\textbf{RL Objective} We denote $\pi_{\boldsymbol{\theta}}$ as the target policy model to be optimized, and $\pi_{\text{ref}}$ as the reference model from which $\pi_{\boldsymbol{\theta}}$ is initialized. $D$ is the query set, $\boldsymbol{x}$ is an input and $\boldsymbol{y}$ is the generated response to $\boldsymbol{x}$. The standard optimization objective of RL is formalized as
\begin{equation}
\label{eq: RL}
\begin{aligned}
\mathcal{O}_{\pi_{\boldsymbol{\theta}}}= \max_{\pi_{\boldsymbol{\theta}}}\mathbb{E}_{\boldsymbol{x}\sim D, \boldsymbol{y}\sim \pi_{\boldsymbol{\theta}}(\cdot | x)}\left[ r(\boldsymbol{x},\boldsymbol{y}) - \beta \mathcal{D}_{\text{KL}}(\pi_{\boldsymbol{\theta}}|| \pi_{ref})\right],
\end{aligned}
\end{equation}
where $r(\boldsymbol{x},\boldsymbol{y})$ represents a reward function to score the response $y$ given $x$, $\mathcal{D}_{\text{KL}}$ is the Kullback–Leibler (KL) divergence loss regularizing the distance between two model distributions. 

\textbf{RLVR} Recently, RLVR~\citep{r1,orz} has emerged as a widely adopted and effective paradigm for enhancing the reasoning capabilities of LLMs. In RLVR, the reward function $r$ is typically chosen as a deterministic verifier $r_v$, such as a rule-based verifier, to evaluate whether the final extracted answer $\boldsymbol{a}\subset \boldsymbol{y}$ matches the ground-truth answer $\boldsymbol{a}^{*}$, and to produce binary feedback (e.g., \{0,1\}). That is,
\begin{equation}
\label{eq: RLVR rewards}
\begin{aligned}
r_{v}(\boldsymbol{x},\boldsymbol{y})= \mathds{1}_{\{\boldsymbol{a}\equiv \boldsymbol{a}^{*}\}}= \begin{cases}
1 & \text{if $\boldsymbol{a}$ is semantically equivalent to $\boldsymbol{a}^{*}$} , \\
0 & \text{otherwise}.
\end{cases}
\end{aligned}
\end{equation}

\textbf{Policy Gradient Method}  Policy Gradient~\citep{rl_introduction} is a widely adopted algorithm to optimize the objective of Eq.~(\ref{eq: RL}), which updates the policy model via the estimated gradient 
\begin{equation}
\label{eq: policy gradient}
\begin{aligned}
\nabla_{\boldsymbol{\theta}}\mathcal{O}_{\pi_{\boldsymbol{\theta}}} = \mathbb{E}_{\boldsymbol{x}\sim D, \boldsymbol{y}\sim \pi_{\boldsymbol{\theta}}(\cdot | x)}\left[\sum_{t=1}^{T} A_{t} \nabla_{\boldsymbol{\theta}} \log \pi_{\boldsymbol{\theta}} (y_{t}|\boldsymbol{x},\boldsymbol{y}_{<t}) \right],
\end{aligned}
\end{equation}
where $A_{t}$ is the \textit{advantage function} measuring the relative value of the action $a_{t}$ (i.e., token $y_t$) compared to the baseline value under state $s_{t}$ (i.e., sequence $(\boldsymbol{x}, \boldsymbol{y}_{<t})$). In practice, $A_{t}$ can be estimated in various ways~\citep{ppo,rloo}. For example, 
Group Relative Policy Optimization~(GRPO)~\citep{deepseekmath} estimates the baseline value as the average reward within a sampled group $\{\boldsymbol{y}^{1},\cdots ,\boldsymbol{y}^{K}\}$ for the same problem, and computes the relative advantage for each token $y_{t}^{i}$ in sequence $\boldsymbol{y}^{i}$ as
\begin{equation}
\label{eq: grpo adv}
\begin{aligned}
A_{t}^{i}= (r_{v}^{i}-\text{mean}(r_{v}^1,\cdots,r_{v}^{K})) / \text{std}(r_{v}^1,\cdots,r_{v}^{K}), \quad r_{v}^{i} = r_{v}(\boldsymbol{x},\boldsymbol{y}^{i}).
\end{aligned}
\end{equation}

\textbf{Implicit Reward} Previous studies~\citep{dpo,rl_by_reward_weighted_regression} have identified that the optimal  solution to the objective Eq.~(\ref{eq: RL}) satisfies that
\begin{equation}
\label{eq: RL solution}
\begin{aligned}
r_{v}(\boldsymbol{x},\boldsymbol{y})=\beta \log [\pi_{\boldsymbol{\theta}}(\boldsymbol{y}|\boldsymbol{x}) / \pi_{ref}(\boldsymbol{y}|\boldsymbol{x})] + \beta \log Z(\boldsymbol{x}),
\end{aligned}
\end{equation}
where $Z(\boldsymbol{x})=\sum_{\boldsymbol{y}} \pi_{ref}(\boldsymbol{y}|\boldsymbol{x}) \exp (\frac{1}{\beta}r_{v}(\boldsymbol{x},\boldsymbol{y}))$ is a partition function. $\beta \log \frac{\pi_{\boldsymbol{\theta}}(\boldsymbol{y}|\boldsymbol{x})}{\pi_{ref}(\boldsymbol{y}|\boldsymbol{x})}$ is termed as the \textit{implicit reward}, which has been used in prior works~\citep{emulated_finetuning,lm_proxy} to analyze the behavioral shift induced by the alignment process.

\subsection{LaSeR: Reinforcement Learning with Last-Token Self-Rewarding}

\subsubsection{Formal Formulation}
In training, ground-truth answers can be reliably used to determine the correctness of solutions. At test time, however, when ground-truth answers are unavailable, the use of verifiers becomes crucial for evaluating solution quality and providing feedback signals. To address this problem, in this work, we explore the promising paradigm of jointly optimizing the self-verification and reasoning capabilities of LLMs within the RLVR framework, thereby enabling them not only to produce high-quality reasoning paths but also to evaluate their own outputs at test time.

According to Eq.~(\ref{eq: RL solution}), as $Z(\boldsymbol{x})$ remains the same for all $\boldsymbol{y}$, a straight-forward idea is to utilize the implicit reward $\beta \log \frac{\pi_{\boldsymbol{\theta}}(\boldsymbol{y}|\boldsymbol{x})}{\pi_{ref}(\boldsymbol{y}|\boldsymbol{x})}$ as the indicator to rank different generations at test time. However, this approach has a critical drawback: the absolute value of the implicit reward is \textbf{length-biased}, since the absolute value of $\beta \log \frac{\pi_{\boldsymbol{\theta}}(\boldsymbol{y}|\boldsymbol{x})}{\pi_{ref}(\boldsymbol{y}|\boldsymbol{x})}=\beta \sum_{i} \log \frac{\pi_{\boldsymbol{\theta}}(y_i|\boldsymbol{x},\boldsymbol{y}_{<i})}{\pi_{ref}(y_{i}|\boldsymbol{x}, ,\boldsymbol{y}_{<i})}$ increases proportionally with the response length. In reasoning tasks, the incorrect solutions are usually longer than the correct solutions~\citep{dont-overthink}, making the implicit reward unreliable in evaluating solution correctness (see Appendix~\ref{appendix: length bias of implicit reward}). Furthermore, disregarding $Z(\boldsymbol{x})$ and directly aligning the implicit reward with the true reasoning reward during training degrades the policy model’s generation ability~\citep{prime}, since a fundamental gap (i.e., $\beta \log Z(\boldsymbol{x})$) exists between the solution to RLVR and that to reward modeling.~\looseness=-1

In this work, we begin by formulating our approach from the RL objective of verification. Given a problem $\boldsymbol{x}$, and a candidate solution $\boldsymbol{y}$, the model is required to produce a verification $\boldsymbol{z}$ to identify the correctness of the solution: $\boldsymbol{z} \sim \pi_{\boldsymbol{\theta}}(\cdot|\boldsymbol{x},\boldsymbol{y})$. Thus, the RL objective of verification can be written as~\looseness=-1
\begin{equation}
\label{eq: RL verification}
\begin{aligned}
\mathcal{V}_{\pi_{\boldsymbol{\theta}}}= \max_{\pi_{\boldsymbol{\theta}}}\mathbb{E}_{\boldsymbol{x}\sim D, \boldsymbol{y}\sim \pi_{g}(\cdot | x),\boldsymbol{z}\sim \pi_{\boldsymbol{\theta}}(\cdot|\boldsymbol{x},\boldsymbol{y})}\left[ \hat{r}(\boldsymbol{x},\boldsymbol{y},\boldsymbol{z}) - \beta_{v} \mathcal{D}_{\text{KL}}(\pi_{\boldsymbol{\theta}}|| \pi_{ref})\right],
\end{aligned}
\end{equation}
where $\pi_{g}$ is the generator to solve the problem (can also be the target model $\pi_{\boldsymbol{\theta}}$ itself in the self-verification setting), $\hat{r}(\boldsymbol{x},\boldsymbol{y},\boldsymbol{z})$ is the verification reward that measures the consistency between the true correctness of $\boldsymbol{y}$ and the verification result of $\boldsymbol{z}$:
\begin{equation}
\label{eq: verification rewards}
\begin{aligned}
\hat{r}(\boldsymbol{x},\boldsymbol{y},\boldsymbol{z})= \begin{cases}
1 & \text{if verification result of $\boldsymbol{z}$ matches the true correctness of $\boldsymbol{y}$} , \\
0 & \text{otherwise}.
\end{cases}
\end{aligned}
\end{equation}
In practice, $\boldsymbol{z}$ can be either a single token—for instance, ``\texttt{Yes}'' or ``\texttt{No}'' to directly indicate whether the solution is verified as correct or incorrect—or a sequence that includes both a chain of thought and the final judgment. In this work, we focus on the former setting and simplify the ground-truth label space to two single tokens $z_{c}$ (e.g., ``\texttt{Yes}'') and $z_{i}$ (e.g., ``\texttt{No}''). That is, the verification reward is simplified to
\begin{equation}
\label{eq: verification rewards simplified}
\begin{aligned}
\hat{r}(\boldsymbol{x},\boldsymbol{y},\boldsymbol{z})= \begin{cases}
1 & (\boldsymbol{z}=z_c \text{ and } r_{v}(\boldsymbol{x},\boldsymbol{y})=1)\text{ or }(\boldsymbol{z}=z_i \text{ and } r_{v}(\boldsymbol{x},\boldsymbol{y})=0) \\
0 & \text{otherwise}.
\end{cases}
\end{aligned}
\end{equation}

Similarly, following from Eq.~(\ref{eq: RL solution}), the close-form solution to Eq.~(\ref{eq: RL verification}) can be written as
\begin{equation}
\label{eq: verification solution}
\begin{small}
\begin{aligned}
\hat{r}(\boldsymbol{x},\boldsymbol{y},\boldsymbol{z})=\beta_{v} \log \frac{\pi_{\boldsymbol{\theta}}(\boldsymbol{z}|\boldsymbol{x},\boldsymbol{y})}{\pi_{ref}(\boldsymbol{z}|\boldsymbol{x},\boldsymbol{y})} + \beta_{v} \log Z(\boldsymbol{x},\boldsymbol{y}),\quad Z(\boldsymbol{x},\boldsymbol{y})=\sum_{\boldsymbol{z}} \pi_{ref}(\boldsymbol{z}|\boldsymbol{x},\boldsymbol{y}) \exp (\frac{1}{\beta_{v}}\hat{r}(\boldsymbol{x},\boldsymbol{y},\boldsymbol{z})).
\end{aligned}
\end{small}
\end{equation}

Now, let's take a closer look at $Z(\boldsymbol{x},\boldsymbol{y})$. First, \textbf{for $\boldsymbol{z}\in \{ z_{c}, z_{i}\}$, $\pi_{ref}(\boldsymbol{z}|\boldsymbol{x},\boldsymbol{y})$ is a extremely small positive value for any problem-solution pair $(\boldsymbol{x}, \boldsymbol{y})$, i.e., $\pi_{ref}(\boldsymbol{z}|\boldsymbol{x},\boldsymbol{y}) \approx 0$, for $\boldsymbol{z}\in \{ z_{c}, z_{i}\}$}. The reason is that the model is not specifically optimized for predicting the next token once it completes the generation and produces the final token (typically the ``\texttt{<EOS>}'' token). We present a numerical analysis to validate this claim in Figure~\ref{fig: small value of log ref}, and we can see \textbf{the value of $\pi_{ref}(z|\boldsymbol{x},\boldsymbol{y})$ is less than $e^{-9}$ for common tokens and even less than $e^{-20}$ for unused special tokens}. Then, we can get that
\begin{equation}
\label{eq: deduction of Z(x,y)}
\begin{small}
\begin{aligned}
Z(\boldsymbol{x},\boldsymbol{y})&=\sum_{\boldsymbol{z}} \pi_{ref}(\boldsymbol{z}|\boldsymbol{x},\boldsymbol{y}) \exp (\frac{1}{\beta_{v}}\hat{r}(\boldsymbol{x},\boldsymbol{y},\boldsymbol{z}))=
\sum_{\boldsymbol{z} \notin \{z_{c},z_{i} \}} \pi_{ref}(\boldsymbol{z}|\boldsymbol{x},\boldsymbol{y}) \exp (\frac{1}{\beta_{v}}\hat{r}(\boldsymbol{x},\boldsymbol{y},\boldsymbol{z}))\\ & + \pi_{ref}(z_{c}|\boldsymbol{x},\boldsymbol{y}) \exp (\frac{1}{\beta_{v}}\hat{r}(\boldsymbol{x},\boldsymbol{y},z_{c}))  + \pi_{ref}(z_{i}|\boldsymbol{x},\boldsymbol{y}) \exp (\frac{1}{\beta_{v}}\hat{r}(\boldsymbol{x},\boldsymbol{y},z_{i})) \\& = (1 -  \pi_{ref}(z_{c}|\boldsymbol{x},\boldsymbol{y}) -\pi_{ref}(z_{i}|\boldsymbol{x},\boldsymbol{y})) \exp (0) + ( \pi_{ref}(z_{c}|\boldsymbol{x},\boldsymbol{y}) +\pi_{ref}(z_{i}|\boldsymbol{x},\boldsymbol{y})) \exp (\frac{1}{\beta_{v}}) \\&
 \approx 1 \times 1 + 0 \times \exp (\frac{1}{\beta_{v}})=1 \implies \log Z(\boldsymbol{x}, \boldsymbol{y}) \approx 0.
\end{aligned}
\end{small}
\end{equation}
The above analysis reveals that, under our formulation, the partition function that cannot be ignored by previous studies~\citep{prime} can be now naturally discarded. Consequently, the optimal solution to Eq.~(\ref{eq: RL verification}) can be \textbf{approximately} reduced to:
\begin{equation}
\label{eq: verification solution simplified}
\begin{aligned}
\hat{r}(\boldsymbol{x},\boldsymbol{y},\boldsymbol{z}) = \beta_{v} \log [\pi_{\boldsymbol{\theta}}(\boldsymbol{z}|\boldsymbol{x},\boldsymbol{y}) / \pi_{ref}(\boldsymbol{z}|\boldsymbol{x},\boldsymbol{y})] .
\end{aligned}
\end{equation}
In particular, the true verification reward when the model verifies a solution as correct is:
\begin{equation}
\label{eq: verification solution simplified yes}
\begin{aligned}
\hat{r}(\boldsymbol{x},\boldsymbol{y},z_{c}) = r_{v}(\boldsymbol{x},\boldsymbol{y})=  \beta_{v} \log [\pi_{\boldsymbol{\theta}}(z_{c}|\boldsymbol{x},\boldsymbol{y}) / \pi_{ref}(z_{c}|\boldsymbol{x},\boldsymbol{y})] .
\end{aligned}
\end{equation}
The first equation is derived from the definition in Eq.~(\ref{eq: verification rewards simplified}). The second equation reveals that \textbf{the true reasoning reward is equal to log-probability ratio of the policy model to the reference model at 
$z_{c}$, scaled by the KL coefficient}. Thus, to optimize the model's verification capability, we do not need to explicitly perform a RLVR procedure. Instead, we can directly optimize the following MSE loss:~\looseness=-1
\begin{equation}
\label{eq: self-rewarding mse loss}
\begin{aligned}
L=\mathbb{E}_{\boldsymbol{x}\sim D, \boldsymbol{y}\sim \pi_{g}(\cdot | x)}\left( \beta_{v} \log [\pi_{\boldsymbol{\theta}}(z_{c}|\boldsymbol{x},\boldsymbol{y})/ \pi_{ref}(z_{c}|\boldsymbol{x},\boldsymbol{y})]  - r_{v}(\boldsymbol{x},\boldsymbol{y})\right)^{2}.
\end{aligned}
\end{equation}

Thus, in the self-verification setting where $\pi_{g}=\pi_{\boldsymbol{\theta}}$, we can directly adds the above loss into the original RLVR loss to jointly optimize the reasoning and self-verification capabilities of the policy model:~\looseness=-1
\begin{equation}
\label{eq: joint loss}
\begin{small}
\begin{aligned}
\mathcal{S}_{\pi_{\boldsymbol{\theta}}}= \max_{\pi_{\boldsymbol{\theta}}}\mathbb{E}_{\boldsymbol{x}\sim D, \boldsymbol{y}\sim \pi_{\boldsymbol{\theta}}(\cdot | x)}\left\{ r_{v}(\boldsymbol{x},\boldsymbol{y}) - \beta \mathcal{D}_{\text{KL}}(\pi_{\boldsymbol{\theta}}(\boldsymbol{y}| \boldsymbol{x})|| \pi_{ref}(\boldsymbol{y}| \boldsymbol{x})) -\alpha \left[\beta_{v} \log \frac{\pi_{\boldsymbol{\theta}}(z_{c}|\boldsymbol{x},\boldsymbol{y})}{\pi_{ref}(z_{c}|\boldsymbol{x},\boldsymbol{y})}  - r_{v}(\boldsymbol{x},\boldsymbol{y})\right]^{2}\right\},
\end{aligned}
\end{small}
\end{equation}
where $\alpha$ is a loss balancing coefficient. We refer the term $r_{s}=\beta_{v} \log \frac{\pi_{\boldsymbol{\theta}}(z_{c}|\boldsymbol{x},\boldsymbol{y})}{\pi_{ref}(z_{c}|\boldsymbol{x},\boldsymbol{y})} $ to the \textbf{last-token self-rewarding score}, since it depends on the log-probability distributions of the last token in $\boldsymbol{y}$.~\looseness=-1

\subsection{Other Techniques}
\label{subsec: other techniques}
Here, we discuss several practical techniques to further simplify and improve the efficiency and effectiveness of the self-rewarding MSE loss introduced above.

\textbf{Simplification of the Log-Probability in the Reference Model} 
As shown in Figure~\ref{fig: small value of log ref}, the quantity 
$\log \pi_{ref}(z_{c}|\boldsymbol{x},\boldsymbol{y})$ remains almost constant, exhibiting only a negligible standard deviation across all $\boldsymbol{x}$ and $\boldsymbol{y}$. Therefore, we can regard it as a pre-calculated constant $c_{ref}$ in calculating the last-token self-rewarding score during both training and inference. This eliminates the need for forwarding 
$\boldsymbol{y}$ through the reference model and thus further enhances efficiency. In specific, $c_{ref}$ is the mean value of $\log \pi_{ref}(z_{c}|\boldsymbol{x},\boldsymbol{y})$ on a small set of pre-generated set of $(\boldsymbol{x},\boldsymbol{y})$. Furthermore, based on the findings in Figure~\ref{fig: small value of log ref}, we select an unused special token as $z_{c}$ to make $\pi_{ref}(z_{c}|\boldsymbol{x},\boldsymbol{y})$ closer to 0 and to further minimize its impact on the approximation of $Z(\boldsymbol{x},\boldsymbol{y}) = 1$ and the stability of training.

\textbf{Self-Rewarding Loss Re-Weighting} During training, the numbers of correct and incorrect solutions are imbalanced, and their ratio dynamically changes. To prevent the last-token self-rewarding score from being biased toward the class with more samples, we apply a class-level loss re-weighting strategy within each optimization step. In each step, we calculate the total numbers of correct and incorrect solutions (identified by the deterministic verifier) for all problems in the current batch as $N_{c}$ and $N_{i}$. Then, we apply the loss re-weighting as 
\begin{equation}
\label{eq: re-weighting}
\begin{small}
\begin{aligned}
l=\frac{1}{N_{c}+N_{i}}\sum_{\boldsymbol{x}}\sum_{\boldsymbol{y}}\left[ w_{c}\mathds{1}_{\{r_{v}(\boldsymbol{x},\boldsymbol{y})=1\}} + w_{i}\mathds{1}_{\{r_{v}(\boldsymbol{x},\boldsymbol{y})=0\}} \right]\left[\beta_{v} \log \pi_{\boldsymbol{\theta}}(z_{c}|\boldsymbol{x},\boldsymbol{y})  - \beta_{v} c_{ref} - r_{v}(\boldsymbol{x},\boldsymbol{y}) \right]^{2},
\end{aligned}
\end{small}
\end{equation}
where $w_{c} = \frac{N_{c} + N_{i}}{2 \times N_{c}}$ and $w_{i} = \frac{N_{c} + N_{i}}{2 \times N_{i}}$ are re-weighting factors. 
This practice achieves a more balanced self-verification capability. We provide empirical validations on this in Appendix~\ref{appendix: effect of class-level re-weighting}. Future work can explore more effective ways to address the issue of imbalanced distribution of solutions.

\textbf{Integration of Verifier-based and Self-Rewarding-based Advantages} The last-token self-rewarding scores can not only be used at test time, but also facilitate the training process through the integration of verifier-based and self-rewarding-based advantages. We believe such practice can help mitigate the issue of misjudgments by rule-based verifiers, which often occur when the format of ground-truth answer is overly complex, and produce more fine-grained rewards. For example, in GRPO, the final advantage can be calculated as:
\begin{equation}
\label{eq: grpo adv combined}
\begin{small}
\begin{aligned}
\hat{A}_{t}^{i} &= (1-\tau)\frac{r_{v}^{i}-\text{mean}(r_{v}^1,\cdots,r_{v}^{K})}{\text{std}(r_{v}^1,\cdots,r_{v}^{K})} + \tau \frac{r_{s}^{i}-\text{mean}(r_{s}^1,\cdots,r_{s}^{K})}{\text{std}(r_{s}^1,\cdots,r_{s}^{K})}  , \\ & \text{where } r_{v}^{i} = r_{v}(\boldsymbol{x},\boldsymbol{y}^{i}) \text{ and } r_{s}^{i} = \beta_{v} \log \pi_{\boldsymbol{\theta}}(z_{c}|\boldsymbol{x},\boldsymbol{y}^{i})  - \beta_{v} c_{ref}.
\end{aligned}
\end{small}
\end{equation}
To stabilize training, 
we adopt a filtering strategy that sets $\tau=0$ for any group whenever the standard deviation $\text{std}(r_{s}^1,\cdots,r_{s}^{K})$ 
 within this group falls below a threshold $T$, which is set to 0.1.

\textbf{Separate Warm-Up of Reasoning and Self-Rewarding Capabilities}
During the initial phase of training, we optimize only the last-token self-rewarding score, without integrating self-rewarding-based advantages into the learning process. After a certain steps when the last-token self-rewarding loss is sufficiently small, we proceed to integrate verifier-based and self-rewarding-based advantages.
Moreover, when training from base (i.e., pre-trained) models, we first perform standard RLVR without incorporating the last-token self-rewarding loss in order to warm up the model’s reasoning capability, followed by a warm-up phase for the self-rewarding capability before the complete integration of verifier-based and self-rewarding-based advantages.  

By combining all the aforementioned techniques, our full algorithm \textbf{Reinforcement Learning with \underline{La}st-Token \underline{Se}lf-\underline{R}ewarding} (\textbf{LaSeR}), is summarized in Algorithm~\ref{alg: laser} and illustrated in Figure~\ref{fig: pipeline}. During the testing phase, once the model generates a solution, we compute the last-token self-rewarding score based on $r_{s}=\beta_{v} \log \pi_{\boldsymbol{\theta}}(z_{c}|\boldsymbol{x},\boldsymbol{y})  - \beta_{v} c_{ref}$. The comparison between this score and 0.5 determines the self-verification outcome of the solution, or the score itself can be further used to perform weighted majority voting.

\begin{algorithm}[t]
\SetAlgoLined
\KwIn{Initial policy model $\pi_{\boldsymbol{\theta}}$, prompts $D$, verifier $r_{v}$, warm-up hyper-parameters $w_{r}$ and $w_{sr}$, coefficient $\beta_{v}$, pre-specified token $z_{c}$, pre-calculated $c_{ref}=\mathbb{E}_{(\boldsymbol{x},\boldsymbol{y})}[\log \pi_{ref}(z_{c}|\boldsymbol{x},\boldsymbol{y})]$}
\For{\text{Step} $s=1,\cdots, S$}{
1. Set $\pi_{old} \leftarrow \pi_{\boldsymbol{\theta}}$;

2. Sample batch prompts $D_{s}$ from $D$; 

3. Generate solutions $\{\boldsymbol{y}^{i} \}_{i=1}^{K}$ for each $\boldsymbol{x} \in D_{s}$; 

4. Calculate verifier-based rewards and advantages (e.g., Eq.~(\ref{eq: grpo adv})), calculate RL loss;

5. If $s \geq w_{r}$, calculate last-token self-rewarding loss based on Eq.~(\ref{eq: re-weighting}) and add it to RL loss;

6. If $s \geq w_{sr}$, calculate self-rewarding-based advantages and perform advantage integration based on Eq.~(\ref{eq: grpo adv combined});

7. Update the policy model $\pi_{\boldsymbol{\theta}}$ using any RL algorithm with integrated loss and advantages;
}
\KwOut{$\pi_{\boldsymbol{\theta}}$}
\caption{\textbf{LaSeR}: Reinforcement Learning with \underline{La}st-Token \underline{Se}lf-\underline{R}ewarding}
\label{alg: laser}
\end{algorithm}

\subsection{Brief Discussion}
\label{subsec: brief discussion}
\textbf{Comparison Between LaSeR and Prior Approaches} Compared with previous methods~\citep{put-value-back-in-rl,trust-but-verify,rl-tango} that requires the policy model to perform separate generations for solutions and verifications, our method directly derives the self-rewarding result from the next-token log-probability of the final solution token. In the RL process, the computation of token log-probabilities is typically carried out after all the generations are completed~\citep{verl}. Therefore, we can directly replace the token id of the first padding token with the token id of the pre-specified token before computing the log-probabilities of the sequences, thereby \textbf{incurring no additional computation cost during training}. \textbf{During inference, our method requires only one more token inference after the solution is completed}, which substantially reduces the computational cost compared to previous methods. We also discuss the potential way to \textbf{further reduce the self-rewarding cost by avoiding any extra token inference} in Section~\ref{subsec: discussion on self-rewarding cost}, which can be an interesting future work.

\textbf{Difference Between Last-Token Self-Rewarding Loss and Supervised Fine-Tuning Loss} An alternative to train the self-verification capability is to optimize the following supervised fine-tuning (SFT) loss by maximizing the next-token probability of the token $z_{c}$ or $z_{i}$ based on the context $(\boldsymbol{x},\boldsymbol{y})$:
\begin{equation}
\label{eq: sft loss}
\begin{small}
\begin{aligned}
L_{SFT}= - \mathbb{E}_{\boldsymbol{x}\sim D, \boldsymbol{y}\sim \pi_{g}(\cdot | x)} \left[ r_{v}(\boldsymbol{x},\boldsymbol{y}) \cdot \log \pi_{\boldsymbol{\theta}} (z_{c}|\boldsymbol{x},\boldsymbol{y})  + (1 - r_{v} (\boldsymbol{x},\boldsymbol{y})) \cdot \log \pi_{\boldsymbol{\theta}} (z_{i}|\boldsymbol{x},\boldsymbol{y}) \right].
\end{aligned}
\end{small}
\end{equation}
The major difference between SFT loss and our last-token self-rewarding loss in Eq.~(\ref{eq: self-rewarding mse loss}) is that the SFT loss drives 
\(\pi_{\boldsymbol{\theta}} (z_{c} | \boldsymbol{x}, \boldsymbol{y})\) to fit $1$ when \(r_{v}(\boldsymbol{x},\boldsymbol{y}) = 1\), which may lead to strong interference with the optimization of reasoning capability. 
In contrast, our loss drives 
\(\pi_{\boldsymbol{\theta}} (z_{c} | \boldsymbol{x}, \boldsymbol{y})\) toward 
\(\exp(1 / \beta_{v}) \cdot \pi_{\text{ref}} (z_{c} | \boldsymbol{x}, \boldsymbol{y})\) for \(r_{v}(\boldsymbol{x},\boldsymbol{y}) = 1.0\). 
When \(\beta_{v}\) is relatively large, 
\(\pi_{\boldsymbol{\theta}} (z_{c} | \boldsymbol{x}, \boldsymbol{y})\) remains still very small, thereby exerting only a negligible influence on the original RLVR optimization 
(e.g., \(\pi_{\boldsymbol{\theta}} (z_{c} | \boldsymbol{x}, \boldsymbol{y}) = e^{-13}\) when 
\(\pi_{\text{ref}} (z_{c} | \boldsymbol{x}, \boldsymbol{y}) = e^{-23}\) and \(\beta_{v} = 0.1\)). 
We provide the empirical comparison in Appendix~\ref{appendix: comparison with SFT loss}.

\section{Experiments}
\subsection{Experimental Settings}

\label{subsec: experimental settings}
\textbf{Base Models and Baselines}
We primarily conduct empirical validations on both LLaMA3.2~\cite{llama3.2} and Qwen2.5~\citep{qwen2.5} architectures, including three base models: OctoThinker-3B-Short-Base~\citep{octothinker} (mid-trained version of LLaMA3.2-3B-Base), Qwen2.5-7B-Base~\citep{qwen2.5} (pre-trained model) and Open-Reasoner-Zero-7B~\citep{orz} (reinforced version of Qwen2.5-7B-Base). In principle, our method can be seamlessly integrated into any RLVR framework, as it only introduces an additional MSE loss term. In this work, we adopt the widely used GRPO~\citep{deepseekmath} as the base algorithm and primarily investigate the effectiveness of applying our method within GRPO, while leaving the exploration on other RL algorithms in the future work.

\textbf{Training and Evaluation Datasets}
We adopt DeepMath-103K~\citep{deepmath}, a large-scale and high-quality mathematical reasoning dataset, for our RL training data. In testing, we evaluate both the reasoning and self-verification performance of each model on five typical math reasoning benchmarks: MATH500~\citep{math}, AMC23~\citep{amc2023}, AIME24~\citep{aime2024}, AIME25~\citep{aime2025}, and OlympiadBench~\citep{olympiadbench}. We also explore the effectiveness of our method in general reasoning tasks beyond math reasoning in Section~\ref{subsec: ood tasks}.

\textbf{Training Settings} The detailed training hyper-parameters of GRPO are put in Appendix~\ref{appendix: training settings}. The prompt template for each model is in Appendix~\ref{appendix: prompt templates}. When applying our method, we set the hyper-parameters $(\beta_{v},\alpha,\tau)=(0.1,0.1, 0.1)$, which are empirically determined based on the observations in Appendix~\ref{appendix: ablation studies on self-rewarding hyper-parameters}. $z_{c}$ is selected as ``\texttt{<vision\_start>}'' for Qwen2.5-7B-Base and Open-Reasoner-Zero-7B, and ``\texttt{<reserved\_special\_token\_0>}'' for OctoThinker-3B-Short-Base. The simplified constant of the reference log-probability, $c_{\text{ref}}$, is $-23.0$ for Qwen2.5-7B-Base and Open-Reasoner-Zero-7B, and $-25.0$ for OctoThinker-3B-Short-Base, as estimated from the results in Figure~\ref{fig: small value of log ref}. The number of reasoning warm-up steps is set to 200 for both Qwen2.5-7B-Base and OctoThinker-3B-Short-Base, and the number of self-rewarding warm-up steps is 200 across all models.~\looseness=-1

\textbf{Evaluation Settings} During generation, we set both the \texttt{temperature} and \texttt{top\_p} to $1.0$ for all models. The \texttt{max\_generation\_len} is 8192. On MATH500 and OlympiadBench, we sample 2 solutions for each problem; whereas on AMC23, AIME24, and AIME25, we sample 32 solutions per problem. We then report the average Pass@1 accuracy of each model on each benchmark. We also evaluate the self-verification performance of each model by computing the self-verification F1 score, defined as the harmonic mean of self-verification accuracy on self-generated correct and incorrect solutions. Baselines perform self-verification based on the prompt in Appendix~\ref{appendix: prompt templates}. Any solution without a final answer is automatically treated as incorrect and excluded from the verification accuracy calculation. Detailed self-verification accuracy results are provided in Appendix~\ref{appendix: self-verification results}.

\subsection{Main Results and Analysis}
\begin{table*}[t]
\caption{Reasoning and self-verification performance of each model on five mathematical reasoning benchmarks. We do not report the results of OctoThinker-based models on AIME24-25, as the number of correct solutions is quite insufficient for a reliable evaluation.}
\label{tab: main results}
\centering
\small
\setlength{\tabcolsep}{4.0pt}
\begin{tabular}{lcccccccccccc}
\toprule
\multirow{3.5}{*}{\begin{tabular}[c]{@{}l@{}}Method \end{tabular}} & \multicolumn{6}{c}{Reasoning Accuracy} & \multicolumn{6}{c}{Self-Verification F1 Score} \\
\cmidrule(lr){2-7}
\cmidrule(lr){8-13}
& \multirow{2}{*}{\begin{tabular}[c]{@{}c@{}}MATH-\\500\end{tabular}} & \multirow{2}{*}{\begin{tabular}[c]{@{}c@{}}AMC-\\23\end{tabular}} & \multirow{2}{*}{\begin{tabular}[c]{@{}c@{}}AIME-\\24\end{tabular}} & \multirow{2}{*}{\begin{tabular}[c]{@{}c@{}}AIME-\\25\end{tabular}} & \multirow{2}{*}{\begin{tabular}[c]{@{}c@{}}Olym.-\\Bench\end{tabular}} & \multirow{2}{*}{\begin{tabular}[c]{@{}c@{}}Avg.\end{tabular}} & \multirow{2}{*}{\begin{tabular}[c]{@{}c@{}}MATH-\\500\end{tabular}} & \multirow{2}{*}{\begin{tabular}[c]{@{}c@{}}AMC-\\23\end{tabular}} & \multirow{2}{*}{\begin{tabular}[c]{@{}c@{}}AIME-\\24\end{tabular}} & \multirow{2}{*}{\begin{tabular}[c]{@{}c@{}}AIME-\\25\end{tabular}} & \multirow{2}{*}{\begin{tabular}[c]{@{}c@{}}Olym.-\\Bench\end{tabular}} & \multirow{2}{*}{\begin{tabular}[c]{@{}c@{}}Avg.\end{tabular}}  \\
\\
\midrule
\multicolumn{13}{l}{\emph{\quad \textbf{OctoThinker-3B-Short-Base}}}   \\
Base & \phantom{0}3.7 & \phantom{0}1.3 & - & - & \phantom{0}1.0 & \phantom{0}2.0 & 22.3 & 11.2 & - & - & 13.7 & 15.7 \\
GRPO & 49.8 & 25.3 & - & - & 17.3& 30.8  &56.9  & 47.3  & -  & - & 48.8  & 51.0 \\
\rowcolor{cyan!10}  LaSeR & \textbf{53.1} & \textbf{27.0}  & - & - & \textbf{18.2} & \textbf{32.8} & 73.6 & 70.2 &- & -& \textbf{73.6} & \textbf{72.5} \\ 
\rowcolor{cyan!10}  ~~ \scriptsize{- \textit{SWA}} &   52.9 & 26.1 & - & - & \textbf{18.2} & 32.4 & \textbf{80.4} & \textbf{70.9} & - & - & 66.0 & 72.4 \\
\midrule
\multicolumn{13}{l}{\emph{\quad \textbf{Qwen2.5-7B-Base}}} \\
Base & 35.8 & 20.6 & \phantom{0}3.5 & \phantom{0}1.6 & 12.3  & 14.8 &  36.4 & 30.8 & 27.6 & 32.9 & 36.9 & 32.9\\
GRPO & 79.9 & 55.9  & \textbf{16.2} &  13.8 & 43.3 & 41.8 & 54.6 & 59.7 & 36.6 & 41.5 & 53.5 & 49.2 \\
\rowcolor{cyan!10}   LaSeR & \textbf{80.2} & 58.1 & 15.4 & \textbf{15.7}   & \textbf{44.1}  & \textbf{42.7}  & \textbf{83.2} & \textbf{82.5} & 79.6  & 74.3  & 78.3 &  79.6 \\
\rowcolor{cyan!10}  ~~ \scriptsize{- \textit{SWA}} &  78.0 & \textbf{58.3} & 15.4 & 12.3 & 41.7 & 41.1  & 79.7 & 80.2  & \textbf{81.3}  & \textbf{74.9}  & \textbf{83.3} & \textbf{79.9} \\
\midrule
\multicolumn{13}{l}{\emph{\quad \textbf{Open-Reasoner-Zero-7B}}} \\
Base & 81.9 & 60.3 & 15.6 & \textbf{15.1} & 46.9  & 44.0 & 26.7 & 51.3 & 45.9 & 55.2 & 37.5  &  43.3\\
GRPO & 83.1 & 61.9& 18.1 & 15.0 & 47.1 & 45.0 & 57.1 & 44.8 & 14.6 & 28.1 & 49.5 & 38.8 \\
\rowcolor{cyan!10}   LaSeR & 82.8 & \textbf{62.7} & \textbf{19.1} & \textbf{15.1} & \textbf{47.8} &  \textbf{45.5}  & 87.2 & \textbf{79.7} & \textbf{64.6} & \textbf{77.7} & \textbf{78.7} & \textbf{77.6} \\
\rowcolor{cyan!10}  ~~ \scriptsize{- \textit{SWA}} & \textbf{83.2} & 62.6 & 19.0 & 14.5  & 47.6 & 45.4 & \textbf{87.5}  &  77.7  & 63.3  & 77.3 & 77.9 & 76.7  \\
\bottomrule
\end{tabular}
\end{table*}

\begin{table*}[t]
\caption{Comparison of verification F1 scores between LaSeR (self-rewarding) and external reward models (Qwen2.5-Math-7B-PRM800K, Qwen2.5-Math-PRM-7B, and Qwen2.5-Math-RM-72B) on responses generated by different policy models.}
\label{tab: prm comparison}
\centering
\small
\begin{tabular}{lcccccc}
\toprule
\multirow{1}{*}{\begin{tabular}[c]{@{}l@{}}Method \end{tabular}} 
& MATH500 & AMC23 & AIME24 & AIME25  & Olym. & Avg. \\
\midrule
\multicolumn{7}{l}{\emph{\quad \textbf{Generator: OctoThinker-3B-Short-LaSeR}}} \\ 
Qwen2.5-Math-7B-PRM800K (7B RM) & 77.0 &  68.9  & - & - & 68.5 & 71.5 \\
Qwen2.5-Math-PRM-7B (7B RM) &  80.9 & 63.5 & - & - & 64.1  & 69.5 \\
Qwen2.5-Math-RM-72B (72B RM) & \textbf{89.2}  & \textbf{71.7} & - & - & 72.9  & \textbf{77.9} \\
\rowcolor{cyan!10}  LaSeR (3B Self-Rewarding) & 73.6 & 70.2 & - & - & \textbf{73.6} &   72.5 \\
\midrule
\multicolumn{7}{l}{\emph{\quad \textbf{Generator: Qwen2.5-7B-Laser}}} \\ 
Qwen2.5-Math-7B-PRM800K (7B RM) & 59.4  & 52.7 & 58.8 & 53.8 & 52.0 & 55.3  \\
Qwen2.5-Math-PRM-7B (7B RM) &  82.5 & 79.2  & 75.1 & 72.3 & 77.8 & 77.4 \\
Qwen2.5-Math-RM-72B (72B RM) & \textbf{87.8} & 80.7 & \textbf{81.3} & \textbf{74.8}  & 75.4  & \textbf{80.0} \\
\rowcolor{cyan!10}  LaSeR (7B Self-Rewarding) & 83.2 & \textbf{82.5} & 79.6 & 74.3 &  \textbf{78.3}& 79.6  \\
\midrule
\multicolumn{7}{l}{\emph{\quad \textbf{Generator: Open-Reasoner-Zero-7B-LaSeR}}} \\ 
Qwen2.5-Math-7B-PRM800K (7B RM) & 56.3 & 42.5  &  51.4 & 50.8 & 38.5 & 47.9 \\
Qwen2.5-Math-PRM-7B (7B RM) & 86.0 & 79.6 &  70.8 & 67.3 & 76.0 & 75.9  \\
Qwen2.5-Math-RM-72B (72B RM) & 86.8 & 79.4 & \textbf{71.0} & 71.4 & 75.5 &  76.8\\
\rowcolor{cyan!10}  LaSeR (7B Self-Rewarding) & \textbf{87.2} & \textbf{79.7} & 64.6 & \textbf{77.7} & \textbf{78.7} & \textbf{77.6} \\
\bottomrule
\end{tabular}
\end{table*}

We put the main results in Table~\ref{tab: main results}.  The key conclusion is that, \textbf{across different model variants, our method not only yields better reasoning performance for the policy model compared with the baseline, but also substantially enhances its self-verification capability by enabling the self-rewarding scores to achieve remarkably high F1 scores}. 

Regarding reasoning performance, applying our algorithm leads to higher accuracy in most settings and consistently yields higher average accuracy on the three base models. We think there are two main reasons for this improvement: (1) First, our method encourages the model to encode its assessment of the overall solution in the final response token, which leads to better confidence calibration. Improved calibration itself can have a positive impact on the model’s learning. (2) Second, by integrating self-rewarding-based advantages into verifier-based advantages, our approach enables more fine-grained advantage estimation, which in turn facilitates more effective learning. For comparison, we report the results without self-rewarding-based advantages (\textit{-SWA}) in Table~\ref{tab: main results}. 

Regarding self-rewarding performance, applying a simple last-token self-rewarding MSE loss substantially enhances the self-rewarding capability of the models, achieving around 80\% self-verification F1 scores. This demonstrates strong self-verification accuracy on both correct and incorrect solutions. To further highlight the self-rewarding capabilities, we display the comparison results of verification F1 scores between LaSeR and three advanced external reward models (Qwen2.5-Math-7B-PRM800K~\citep{lessons_of_prm}, Qwen2.5-Math-PRM-7B~\citep{lessons_of_prm}, and Qwen2.5-Math-RM-72B~\citep{qwen2.5-math}) on evaluating the solutions generated by the different reinforced models, i.e., OctoThinker-3B-Short-LaSeR, Qwen2.5-7B-LaSeR, and Open-Reasoner-Zero-7B-LaSeR. The results in Table~\ref{tab: prm comparison} show that LaSeR outperforms equally sized state-of-the-art external verifiers in assessing the model’s own solutions, and even matches the verification performance of a 72B reward model, demonstrating its non-trivial effectiveness in enhancing self-rewarding capability. 
Also, our method requires one additional token inference only to compute the self-rewarding scores for enabling the policy model to function simultaneously as both the generator and reward model, which is highly efficient and practical.

\subsection{Inference-Time Scaling Results}

\begin{figure*}[t]
  \centering
  \subfigure[Results of OT-3B-based models on MATH500 \label{fig: maj on math500 ot}]{\includegraphics[width=0.32\textwidth]{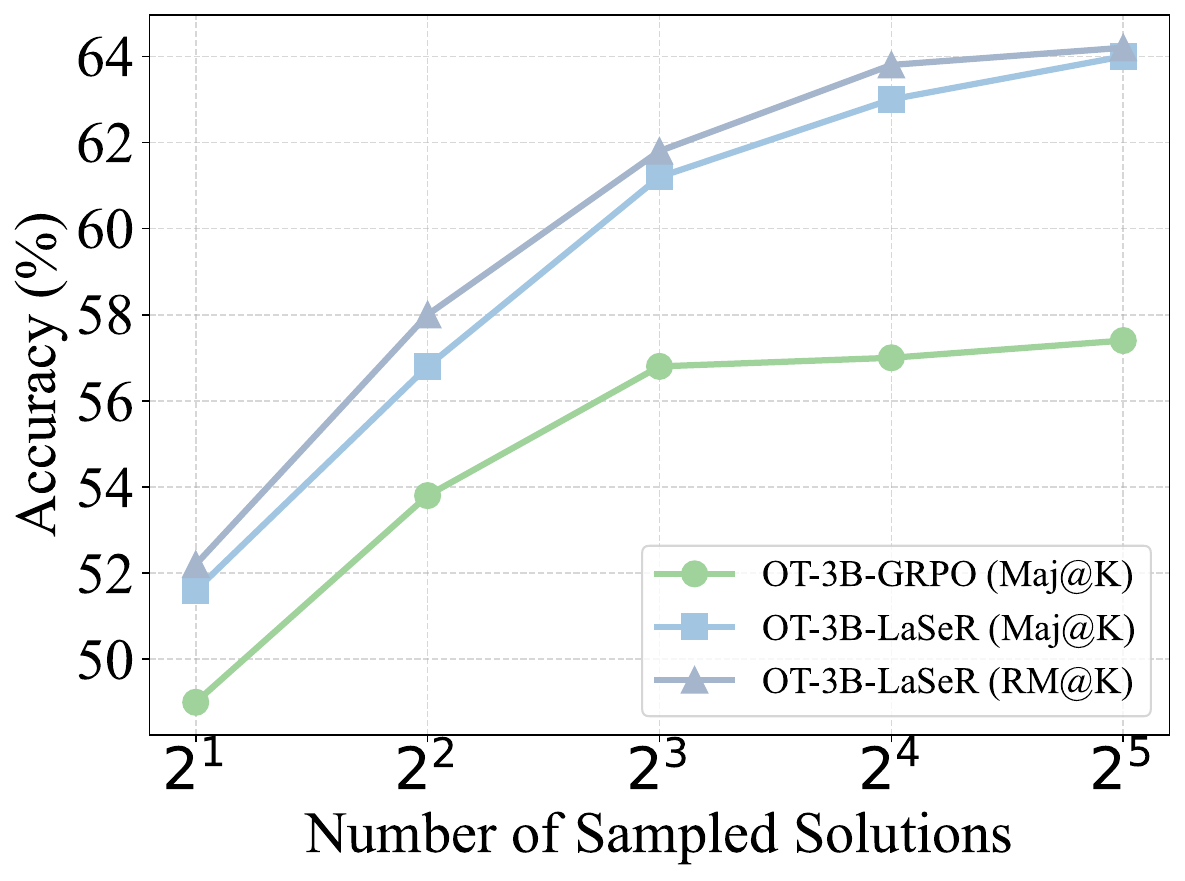}
  }
  \hfill
  \subfigure[Results of Qwen2.5-7B-based models on MATH500 \label{fig: maj on math500 qwen25}]{\includegraphics[width=0.32\textwidth]{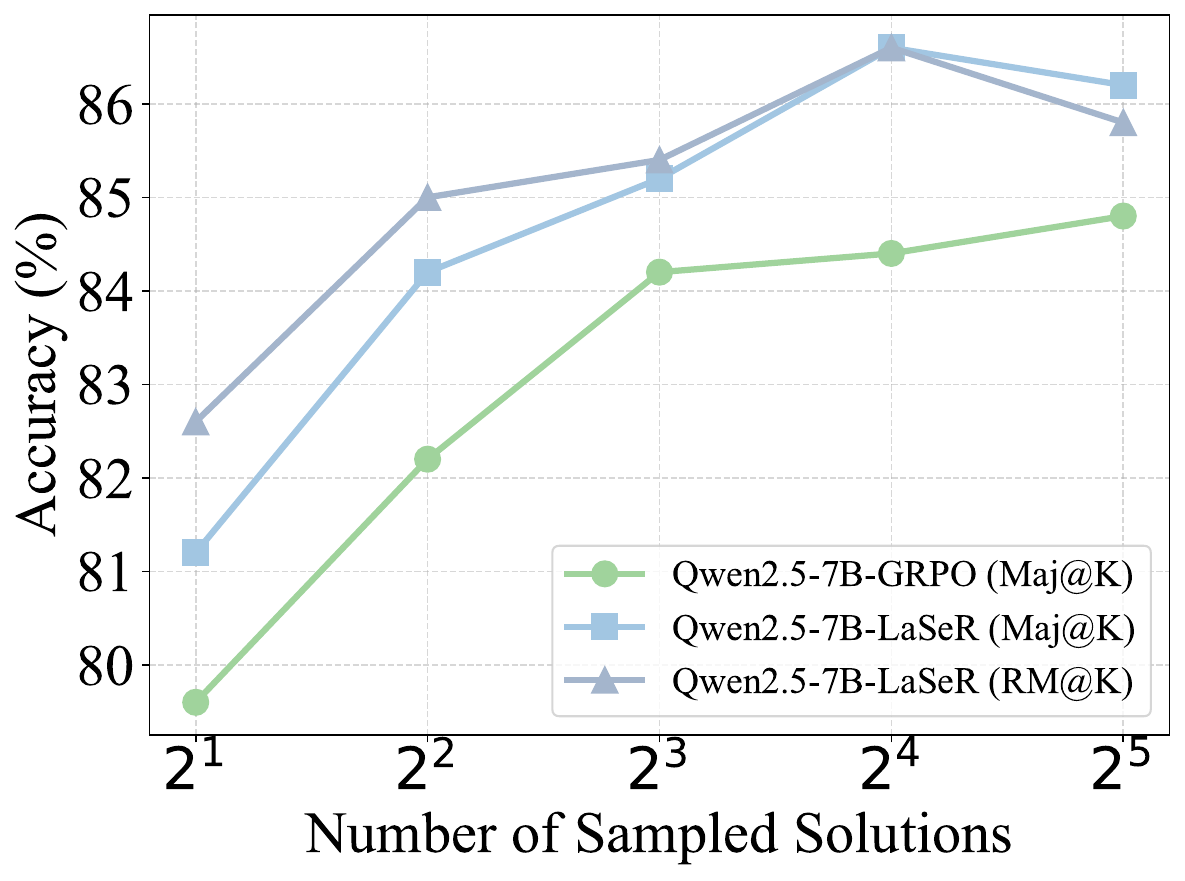}
  }
  \hfill
  \subfigure[Results of ORZ-7B-based models on MATH500  \label{fig: maj on math500 orz}]{\includegraphics[width=0.32\textwidth]{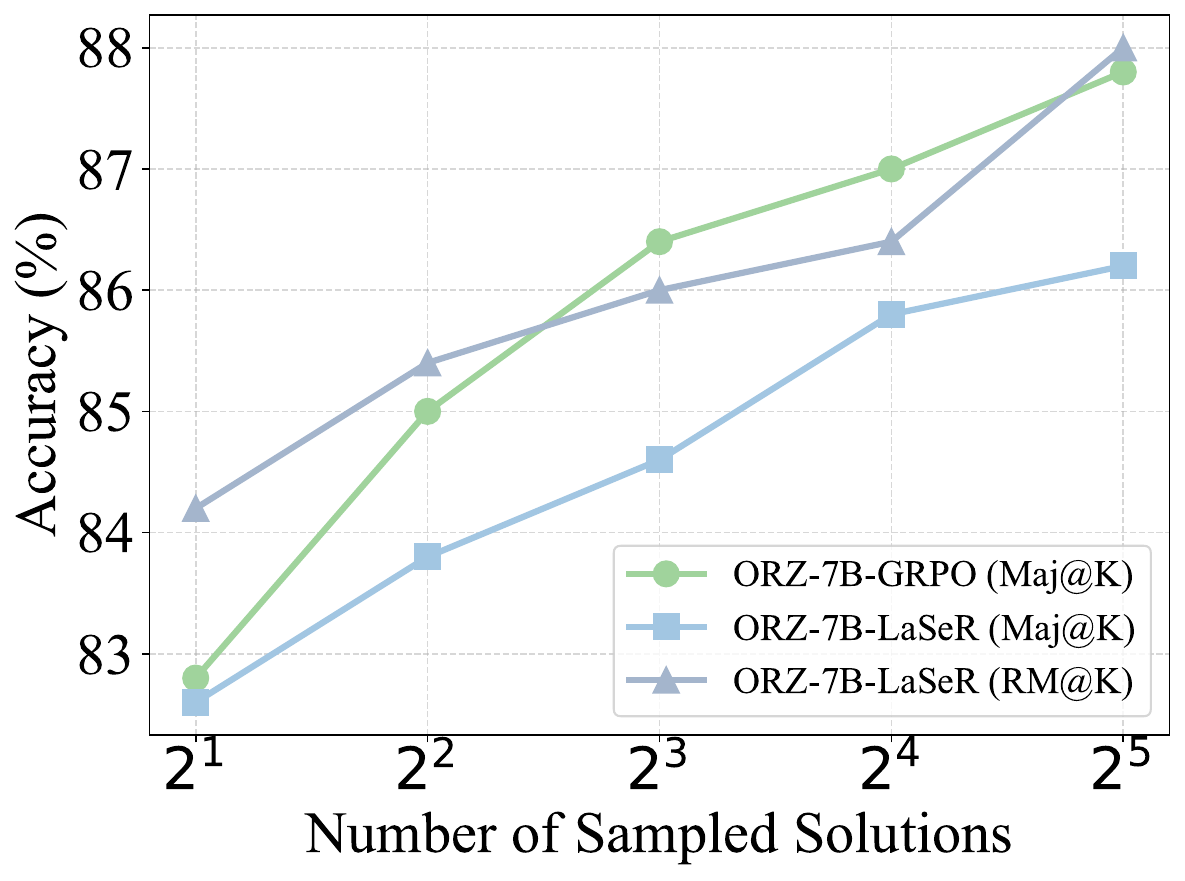}
  }
  \hfill
   \subfigure[Results of OT-3B-based models on OlympiadBench \label{fig: maj on olym ot}]{\includegraphics[width=0.31\textwidth]{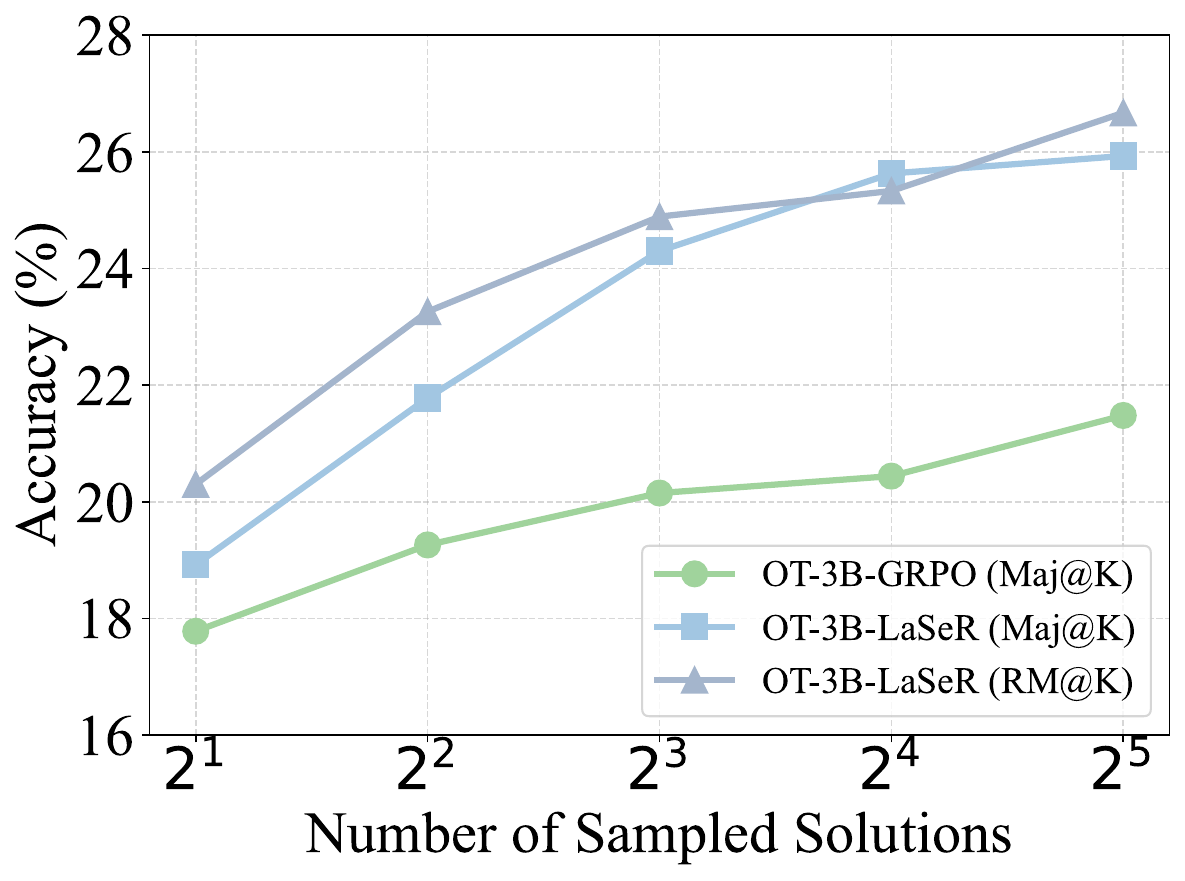}
  }
  \hfill
  \subfigure[Results of Qwen2.5-7B-based models on OlympiadBench \label{fig: maj on olym qwen25}]{\includegraphics[width=0.31\textwidth]{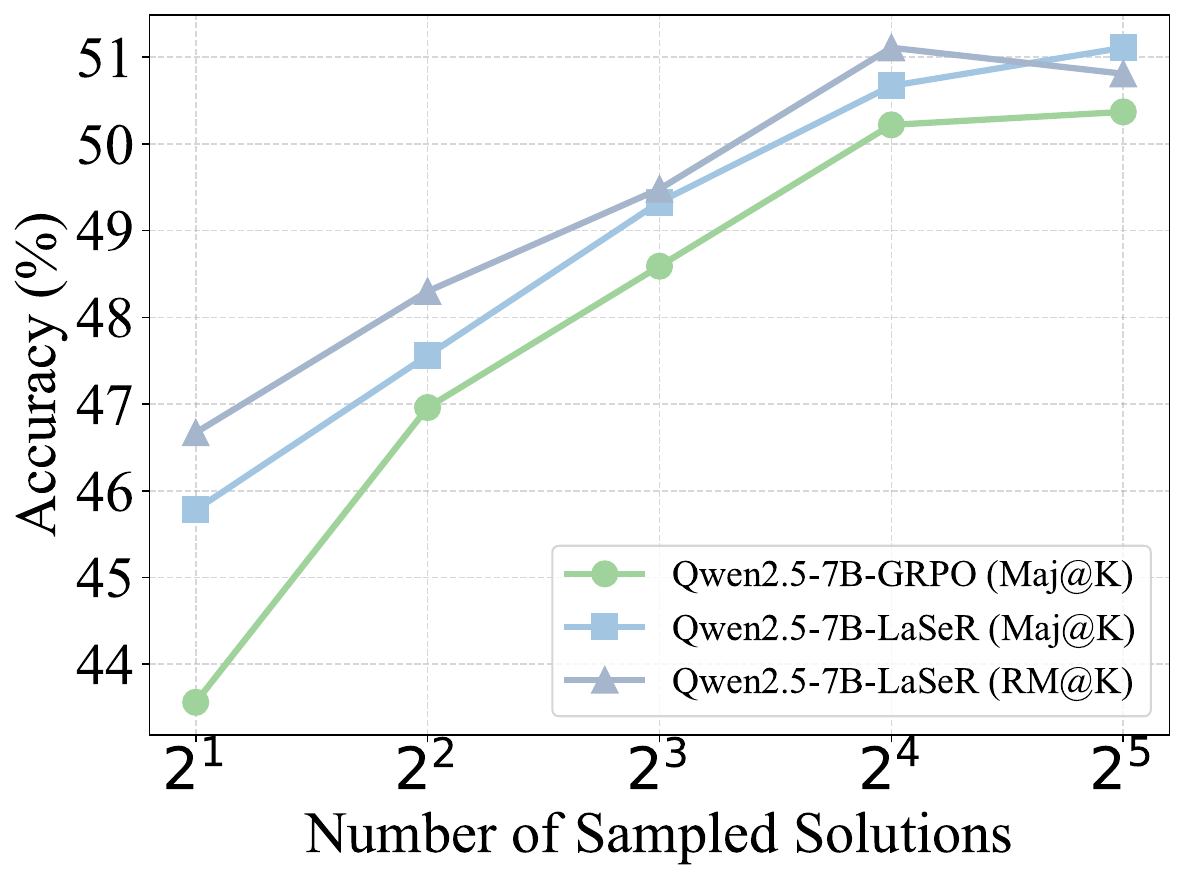}
  }
  \hfill
  \subfigure[Results of ORZ-7B-based models on  OlympiadBench \label{fig: maj on olym orz}]{\includegraphics[width=0.31\textwidth]{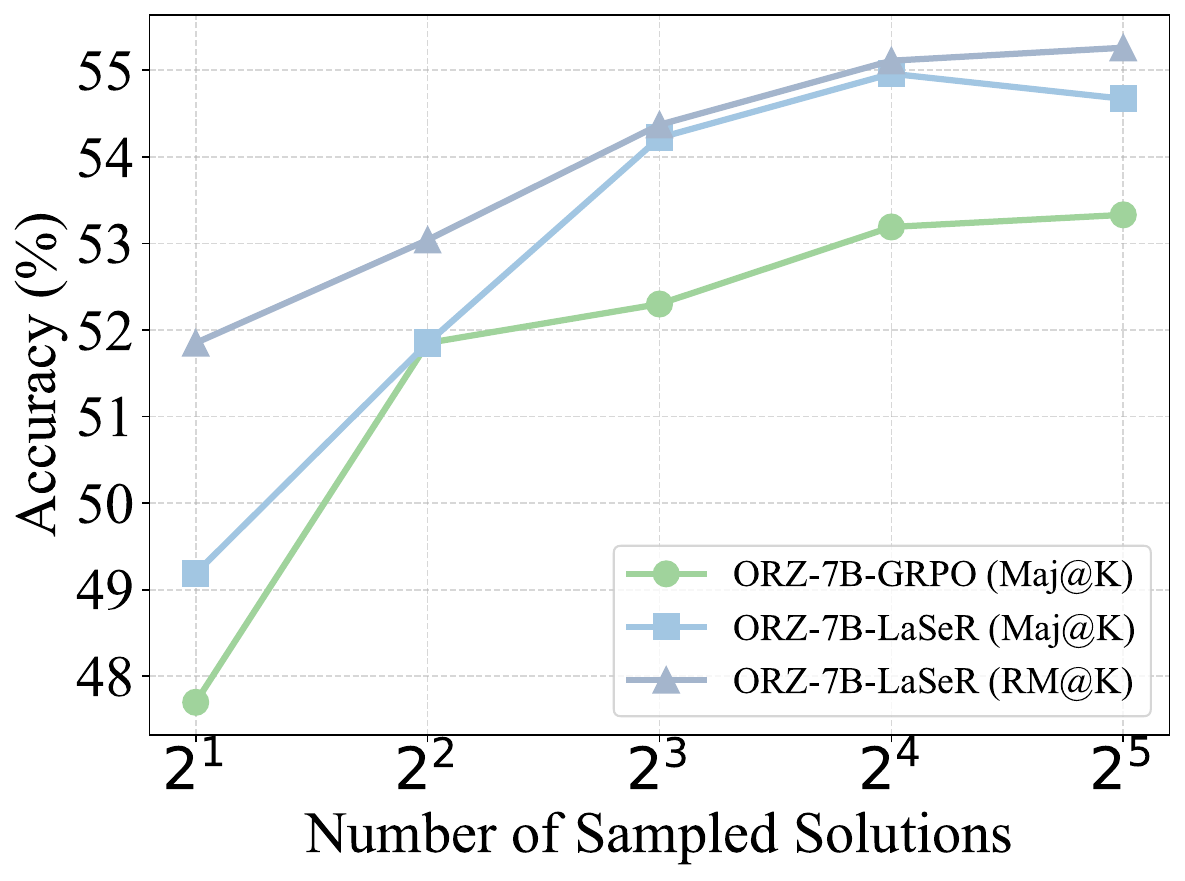}
  }
  \hfill
  \caption{The majority voting (Maj@K) and weighted majority voting (RM@K) results on MATH500 and OlympiadBench.
  }
  \label{fig: maj@k}
\end{figure*}

Here, we explore the effectiveness of self-rewarding in the inference-time scaling via weighted majority voting. We compare majority voting results with (RM@K) and without (Maj@K) weighting by the last-token self-rewarding scores, on MATH500 and OlympiadBench.  
The results are shown in Figure~\ref{fig: maj@k}. 
We denote the three base models by ``OT-3B'', ``Qwen2.5-7B'', and ``ORZ-7B''. The suffixes ``-GRPO'' and ``-LaSeR'' indicate the variants trained with GRPO and our method LaSeR, respectively. 
The results show that \textbf{the optimized self-rewarding capability of the model is highly effective on improving its own inference-time scaling performance}.

\section{Analysis and Discussion}

\begin{table*}[t]
\caption{Comparison of reasoning and self-verification performance with and without reference log-probability simplification in our method. Based model is Open-Reasoner-Zero-7B.}
\label{tab: ref log simplification}
\centering
\small
\setlength{\tabcolsep}{3.5pt}
\begin{tabular}{lcccccccccccc}
\toprule
\multirow{3.5}{*}{\begin{tabular}[c]{@{}l@{}}Method \end{tabular}} & \multicolumn{6}{c}{Reasoning Accuracy} & \multicolumn{6}{c}{Self-Verification F1 Score} \\
\cmidrule(lr){2-7}
\cmidrule(lr){8-13}
& \multirow{2}{*}{\begin{tabular}[c]{@{}c@{}}MATH-\\500\end{tabular}} & \multirow{2}{*}{\begin{tabular}[c]{@{}c@{}}AMC-\\23\end{tabular}} & \multirow{2}{*}{\begin{tabular}[c]{@{}c@{}}AIME-\\24\end{tabular}} & \multirow{2}{*}{\begin{tabular}[c]{@{}c@{}}AIME-\\25\end{tabular}} & \multirow{2}{*}{\begin{tabular}[c]{@{}c@{}}Olym.-\\Bench\end{tabular}} & \multirow{2}{*}{\begin{tabular}[c]{@{}c@{}}Avg.\end{tabular}} & \multirow{2}{*}{\begin{tabular}[c]{@{}c@{}}MATH-\\500\end{tabular}} & \multirow{2}{*}{\begin{tabular}[c]{@{}c@{}}AMC-\\23\end{tabular}} & \multirow{2}{*}{\begin{tabular}[c]{@{}c@{}}AIME-\\24\end{tabular}} & \multirow{2}{*}{\begin{tabular}[c]{@{}c@{}}AIME-\\25\end{tabular}} & \multirow{2}{*}{\begin{tabular}[c]{@{}c@{}}Olym.-\\Bench\end{tabular}} & \multirow{2}{*}{\begin{tabular}[c]{@{}c@{}}Avg.\end{tabular}}  \\
\\
\midrule
w/ Simpl. & 82.5 & 61.6 & 18.8 & 16.2  & 46.5 & 45.1 & 82.3  &  79.3  & 77.9 & 79.2 & 78.4 & 79.4  \\
w/o Simpl. & 81.0 & 61.2 & 17.3 & 17.3 & 48.3 & 45.0  & 81.8 & 79.2 & 79.0 & 78.9 & 77.5 & 79.3\\
\bottomrule
\end{tabular}
\end{table*}

\subsection{The Impact of Simplifying The Reference Log-Probabilities to A Constant}
\label{subsec: log ref simplification}
As discussed in Section~\ref{subsec: other techniques}, 
we approximate the log-probability of the pre-specified token under the reference model, $\log \pi_{ref}(z_{c}|\boldsymbol{x},\boldsymbol{y})$, by its mean computed over a small sample set when calculating the last-token self-rewarding scores. Here, we empirically validate this practice by conducting comparison experiments on Open-Reasoner-Zero-7B, with and without reference log-probability simplification in our method. We evaluate 
the checkpoint after 200 optimization steps in each setting (corresponding to the last checkpoint before advantage integration). The results are reported in Table~\ref{tab: ref log simplification}. As shown, \textbf{the simplification does not affect the optimization of reasoning and self-rewarding capabilities}, since the performance under the two settings remains comparable. However, it effectively reduces the computational cost of calculating the last-token self-rewarding value by half.

\subsection{The Generalizability of LaSeR to General Reasoning Domain}
\label{subsec: ood tasks}

\begin{figure*}[t]
  \centering
  \subfigure[Evaluation accuracy on MMLU-Pro and GPQA-Diamond \label{fig: acc on general reasoning tasks}]{\includegraphics[width=0.32\textwidth]{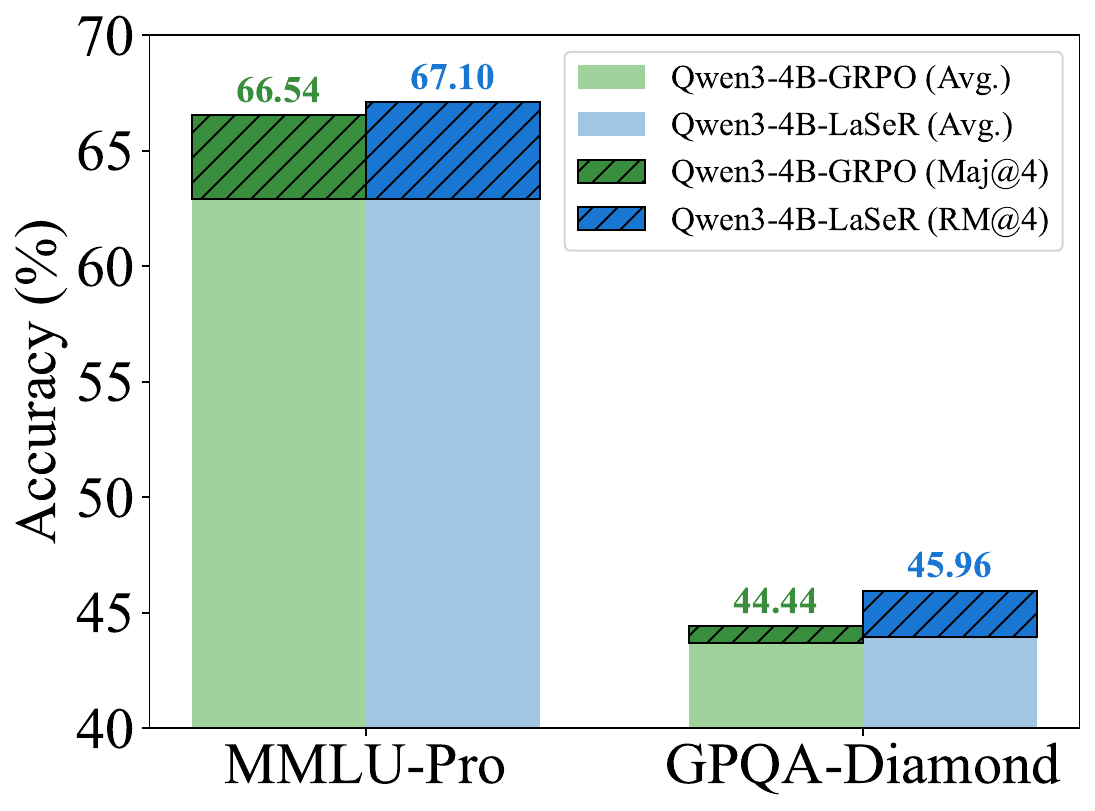}
  }
  \hfill
  \subfigure[Self-rewarding score distribution on MMLU-Pro \label{fig: score distribution on mmlu}]{\includegraphics[width=0.32\textwidth]{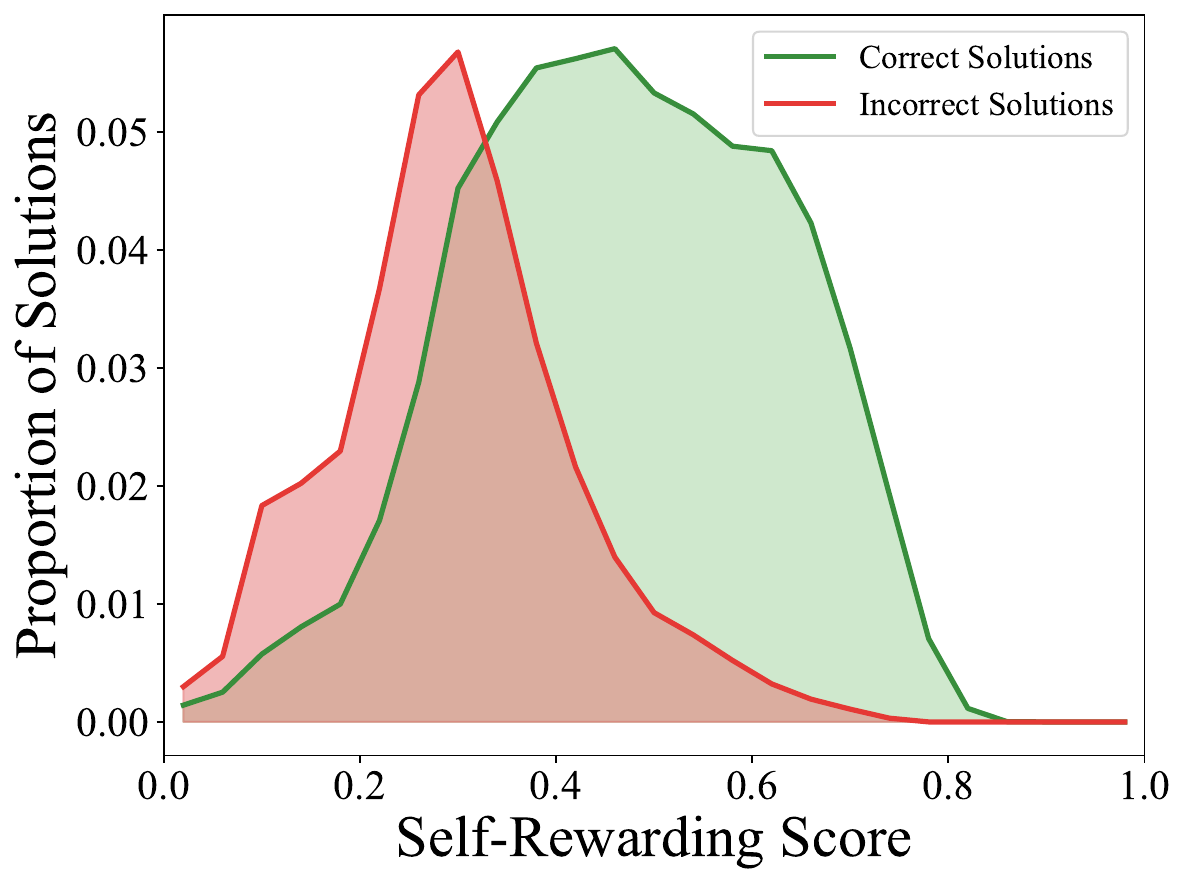}
  }
  \hfill
  \subfigure[Self-rewarding score distribution on GPQA-Diamond \label{fig: score distribution on gpqa}]{\includegraphics[width=0.32\textwidth]{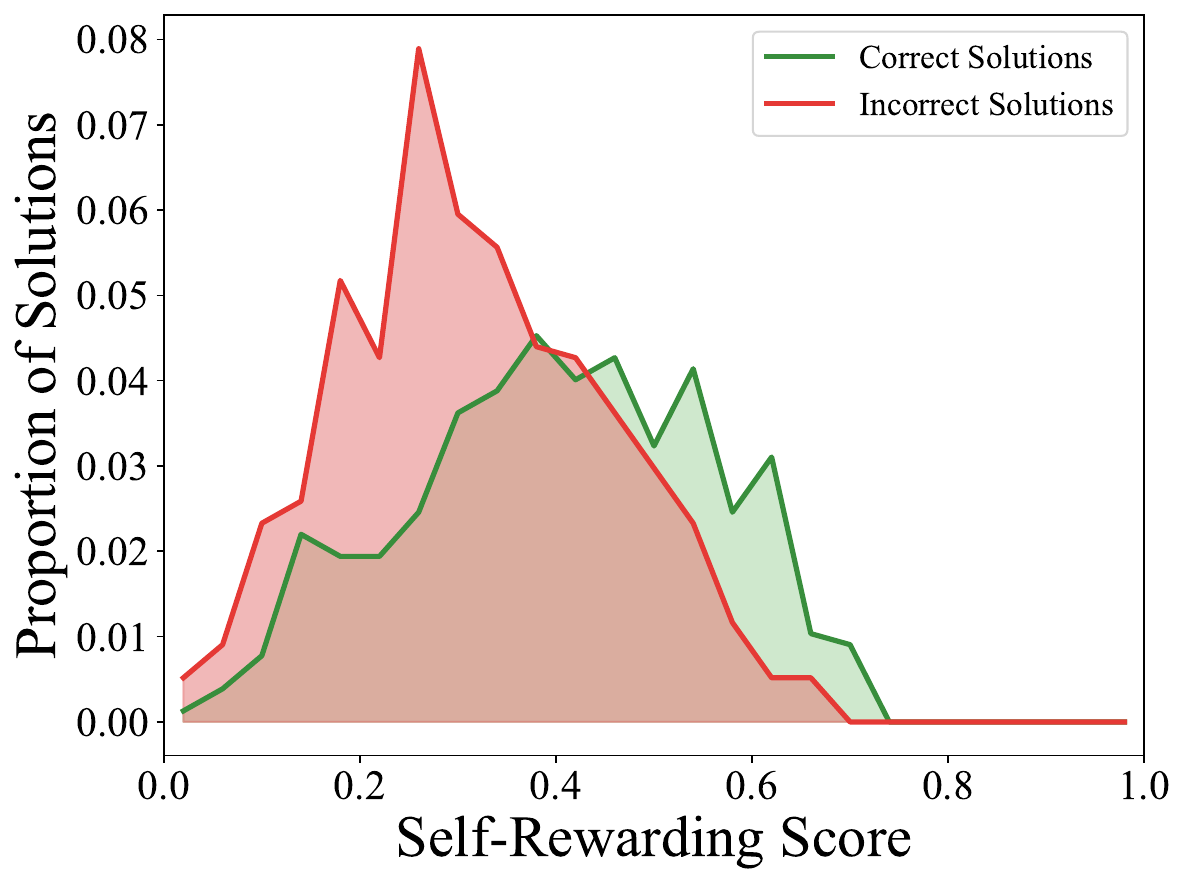}
  }
  \hfill
  \caption{The generalizability of LaSeR on general reasoning tasks.
  }
  \label{fig: results on general reasoning tasks}
\end{figure*}

We conduct additional experiments to explore the generalizability of our method to general reasoning domain. We use a filtered version~\citep{rlpr} of WebInstruct-verified dataset~\citep{general-reasoner}, and conduct RL experiments on Qwen3-4B-Base~\citep{qwen3}. We use the \texttt{general-verifier-1.5B} model from~\citet{general-reasoner} as the model-based verifier and adopt GRPO as the RL algorithm. For our method, we do not perform the advantage integration strategy here. The reason is that we observe the self-verification F1 score of our method during training is relatively low in the general reasoning setting (only between 65\% and 70\%, and the self-rewarding score distributions in the test sets shown in Figure~\ref{fig: score distribution on mmlu} and Figure~\ref{fig: score distribution on gpqa} also reveal this phenomenon). This leads to large noise in the self-rewarding-based advantage estimation, and consequently, the integration of self-rewarding-based advantages results in performance degradation. 
After training, we conduct evaluations on two general reasoning benchmarks: MMLU-Pro~\citep{mmlu-pro} and GPQA-Diamond~\citep{gpqa}. We sample 4 solutions per problem on each dataset for each model, and calculate both the average accuracy and the (weighted) majority voting accuracy. Detailed training and evaluation settings are in Appendix~\ref{appendix: settings on general reasoning tasks}. 

We display the evaluation accuracy in Figure~\ref{fig: acc on general reasoning tasks}, and additionally, we display the self-rewarding score distributions on two datasets in Figure~\ref{fig: score distribution on mmlu} and Figure~\ref{fig: score distribution on gpqa} for reference. First, we observe that jointly optimizing the self-rewarding capability does not impact the model's general reasoning ability, allowing the policy model to achieve comparable average reasoning accuracy to the baseline. However, as mentioned above, the optimized self-rewarding score on general reasoning tasks does not achieve the high accuracy seen in math reasoning tasks. We can see that the self-rewarding score distributions for correct and incorrect solutions on MMLU-Pro exhibit certain overlap, and the distinction further diminishes on the more challenging benchmark GPQA-Diamond. We speculate that two factors may contribute to this: (1) The model's general reasoning ability is inherently weaker than its math reasoning ability, which limits the upper bound of its self-rewarding capabilities in the general reasoning domain. (2) The model-based verifier used in the experiment (\texttt{general-verifier-1.5B}) has limited verification ability, resulting in high noise in the reasoning rewards, which in turn affects the optimization of the self-rewarding capability. A promising direction for future work is to further explore and unlock the full potential of our method in the general reasoning domain. Though not perfect, the optimized self-rewarding scores can still provide useful signals during inference time, leading to better weighted majority voting results.

\subsection{Further Reduction or Increase of Self-Rewarding Cost}
\label{subsec: discussion on self-rewarding cost}
In this section, we discuss two additional variants of LaSeR for future work. In the current method, we calculate the last-token self-rewarding score based on the next-token log-probability distribution of the ``\texttt{<EOS>}'' token, requiring one additional token inference compared with standard inference. One potential way to further reduce the inference cost of LaSeR is to derive the last-token self-rewarding score directly from the predicted log-probability of pre-specified token $z_{c}$ at the ``\texttt{<EOS>}'' token position. Specifically, let $y_{T}$ denote the ``\texttt{<EOS>}'' token in $\boldsymbol{y}$. Then, the reduced last-token self-rewarding score can be defined as $r_{s} = \beta_{v} \log \pi_{\boldsymbol{\theta}}(z_{c}|\boldsymbol{x},\boldsymbol{y}_{<T})  - \beta_{v} c_{ref}$, as we have observed that $\pi_{\text{ref}}(z_{c} | \boldsymbol{x}, \boldsymbol{y}_{<T})$ remains nearly constant across $(\boldsymbol{x}, \boldsymbol{y})$ (e.g., approximately $e^{-28}$ for Qwen2.5-7B-Base). In this case, \textbf{we can achieve ideally zero additional inference cost for self-rewarding compared with standard generation} by directly calculating the self-rewarding score from the log-probability distribution at the ``\texttt{<EOS>}'' token position, without requiring any extra token inference. In theory, this works because setting a relatively large $\beta_{v}$ still yields a small value of $\pi_{\boldsymbol{\theta}}(z_{c} | \boldsymbol{x}, \boldsymbol{y}_{<T})$ (e.g., $\pi_{\boldsymbol{\theta}}(z_{c} | \boldsymbol{x}, \boldsymbol{y}_{<T}) = e^{-18}$ when $\beta_{v} = 0.1$ and $c_{\text{ref}} = -28$), thereby allowing $\pi_{\boldsymbol{\theta}}(\texttt{<EOS>} | \boldsymbol{x}, \boldsymbol{y}_{<T})$ to still dominate the probability mass. However, although the probability is very low, we observe that the generator may still select $z_{c}$ at the end of the sequence in few cases during training, which can adversely affect training stability as the generator continues to generate after $z_{c}$. One straight-forward solution may be to set the sampling hyper-parameter $top\_p$ to a value less than 1.0. Future work can further investigate advanced strategies to make the above adjustment more principled and robust.

While reducing the self-rewarding cost improves efficiency, an alternative is \textbf{to increase the inference cost in exchange for stronger self-rewarding capability}. That is, instead of computing the self-rewarding score based on the log-probability distribution of a single token only, we can increase the number of additional inference tokens by calculating it over $M$ tokens as $r_{s} = \beta_{v} \sum_{m=1}^{M} \log \pi_{\boldsymbol{\theta}}(z_{c}|\boldsymbol{x},\boldsymbol{y},\underbrace{z_{c} ,\cdots,z_{c}}_{m-1\ \text{times}}))  - M \beta_{v} c_{ref}$. It is a promising direction for future research to explore whether increasing the number of additional inference tokens can yield positive inference-time scaling effect for latent self-rewarding capability.

\section{Conclusion}
In this work, we propose \textbf{LaSeR}, a lightweight and effective algorithm that jointly optimizes the reasoning and self-rewarding capabilities of LLMs. By deriving the closed-form solution to the RL objective of verification, we uncover a concise yet intriguing formula: the true reasoning reward provided by the verifier is equal to the last-token self-rewarding score produced by the model. This self-rewarding score depends on the model's next-token log-probability for a pre-specified token at the final response token, a pre-calculated constant, and the KL coefficient. Based on this insight, our method simply adds a MSE loss between the verifier-based reasoning rewards and the corresponding last-token self-rewarding scores into the standard RLVR process. The optimized self-rewarding scores can not only be incorporated back into the RL process to further enhance training, but also achieve high verification accuracy at test time, thereby improving solution ranking and selection.

\bibliography{iclr2025_conference}

\begin{thebibliography}{68}
\providecommand{\natexlab}[1]{#1}
\providecommand{\url}[1]{\texttt{#1}}
\expandafter\ifx\csname urlstyle\endcsname\relax
  \providecommand{\doi}[1]{doi: #1}\else
  \providecommand{\doi}{doi: \begingroup \urlstyle{rm}\Url}\fi

\bibitem[Achiam et~al.(2023)Achiam, Adler, Agarwal, Ahmad, Akkaya, Aleman, Almeida, Altenschmidt, Altman, Anadkat, et~al.]{gpt4}
Josh Achiam, Steven Adler, Sandhini Agarwal, Lama Ahmad, Ilge Akkaya, Florencia~Leoni Aleman, Diogo Almeida, Janko Altenschmidt, Sam Altman, Shyamal Anadkat, et~al.
\newblock Gpt-4 technical report.
\newblock \emph{arXiv preprint arXiv:2303.08774}, 2023.

\bibitem[Ahmadian et~al.(2024)Ahmadian, Cremer, Gall{\'e}, Fadaee, Kreutzer, Pietquin, {\"U}st{\"u}n, and Hooker]{rloo}
Arash Ahmadian, Chris Cremer, Matthias Gall{\'e}, Marzieh Fadaee, Julia Kreutzer, Olivier Pietquin, Ahmet {\"U}st{\"u}n, and Sara Hooker.
\newblock Back to basics: Revisiting reinforce style optimization for learning from human feedback in llms.
\newblock \emph{arXiv preprint arXiv:2402.14740}, 2024.

\bibitem[AI-MO(2024{\natexlab{a}})]{aime2024}
AI-MO.
\newblock Aime 2024.
\newblock \url{https://huggingface.co/datasets/AI-MO/aimo-validation-aime}, 2024{\natexlab{a}}.

\bibitem[AI-MO(2024{\natexlab{b}})]{amc2023}
AI-MO.
\newblock Amc 2023.
\newblock \url{https://huggingface.co/datasets/AI-MO/aimo-validation-amc}, 2024{\natexlab{b}}.

\bibitem[Cobbe et~al.(2021)Cobbe, Kosaraju, Bavarian, Chen, Jun, Kaiser, Plappert, Tworek, Hilton, Nakano, et~al.]{gsm8k}
Karl Cobbe, Vineet Kosaraju, Mohammad Bavarian, Mark Chen, Heewoo Jun, Lukasz Kaiser, Matthias Plappert, Jerry Tworek, Jacob Hilton, Reiichiro Nakano, et~al.
\newblock Training verifiers to solve math word problems.
\newblock \emph{arXiv preprint arXiv:2110.14168}, 2021.

\bibitem[Cui et~al.(2025)Cui, Yuan, Wang, Wang, Li, He, Fan, Yu, Xu, Chen, et~al.]{prime}
Ganqu Cui, Lifan Yuan, Zefan Wang, Hanbin Wang, Wendi Li, Bingxiang He, Yuchen Fan, Tianyu Yu, Qixin Xu, Weize Chen, et~al.
\newblock Process reinforcement through implicit rewards.
\newblock \emph{arXiv preprint arXiv:2502.01456}, 2025.

\bibitem[Dong et~al.(2025)Dong, Mao, Ma, Bao, Chen, Wang, Chen, Du, Wang, Zhang, et~al.]{arpo}
Guanting Dong, Hangyu Mao, Kai Ma, Licheng Bao, Yifei Chen, Zhongyuan Wang, Zhongxia Chen, Jiazhen Du, Huiyang Wang, Fuzheng Zhang, et~al.
\newblock Agentic reinforced policy optimization.
\newblock \emph{arXiv preprint arXiv:2507.19849}, 2025.

\bibitem[Fan et~al.(2025)Fan, Wang, and Liu]{megascience}
Run-Ze Fan, Zengzhi Wang, and Pengfei Liu.
\newblock Megascience: Pushing the frontiers of post-training datasets for science reasoning.
\newblock \emph{arXiv preprint arXiv:2507.16812}, 2025.

\bibitem[Gao et~al.(2024)Gao, Cai, Xu, Wang, Zheng, Lin, Lu, Liu, Zhou, Xiao, et~al.]{llm-critic-catch-math-bugs}
Bofei Gao, Zefan Cai, Runxin Xu, Peiyi Wang, Ce~Zheng, Runji Lin, Keming Lu, Dayiheng Liu, Chang Zhou, Wen Xiao, et~al.
\newblock Llm critics help catch bugs in mathematics: Towards a better mathematical verifier with natural language feedback.
\newblock \emph{arXiv preprint arXiv:2406.14024}, 2024.

\bibitem[Gao et~al.(2023)Gao, Schulman, and Hilton]{reward_hacking}
Leo Gao, John Schulman, and Jacob Hilton.
\newblock Scaling laws for reward model overoptimization.
\newblock In \emph{International Conference on Machine Learning}, pp.\  10835--10866. PMLR, 2023.

\bibitem[Guo et~al.(2025)Guo, Yang, Zhang, Song, Zhang, Xu, Zhu, Ma, Wang, Bi, et~al.]{r1}
Daya Guo, Dejian Yang, Haowei Zhang, Junxiao Song, Ruoyu Zhang, Runxin Xu, Qihao Zhu, Shirong Ma, Peiyi Wang, Xiao Bi, et~al.
\newblock Deepseek-r1: Incentivizing reasoning capability in llms via reinforcement learning.
\newblock \emph{arXiv preprint arXiv:2501.12948}, 2025.

\bibitem[Hassid et~al.(2025)Hassid, Synnaeve, Adi, and Schwartz]{dont-overthink}
Michael Hassid, Gabriel Synnaeve, Yossi Adi, and Roy Schwartz.
\newblock Don't overthink it. preferring shorter thinking chains for improved llm reasoning.
\newblock \emph{arXiv preprint arXiv:2505.17813}, 2025.

\bibitem[He et~al.(2024)He, Luo, Bai, Hu, Thai, Shen, Hu, Han, Huang, Zhang, et~al.]{olympiadbench}
Chaoqun He, Renjie Luo, Yuzhuo Bai, Shengding Hu, Zhen Thai, Junhao Shen, Jinyi Hu, Xu~Han, Yujie Huang, Yuxiang Zhang, et~al.
\newblock Olympiadbench: A challenging benchmark for promoting agi with olympiad-level bilingual multimodal scientific problems.
\newblock In \emph{Proceedings of the 62nd Annual Meeting of the Association for Computational Linguistics (Volume 1: Long Papers)}, pp.\  3828--3850, 2024.

\bibitem[He et~al.(2025)He, Liang, Xu, Liu, Chen, Wang, Song, Yu, Liang, Wang, et~al.]{deepmath}
Zhiwei He, Tian Liang, Jiahao Xu, Qiuzhi Liu, Xingyu Chen, Yue Wang, Linfeng Song, Dian Yu, Zhenwen Liang, Wenxuan Wang, et~al.
\newblock Deepmath-103k: A large-scale, challenging, decontaminated, and verifiable mathematical dataset for advancing reasoning.
\newblock \emph{arXiv preprint arXiv:2504.11456}, 2025.

\bibitem[Hendrycks et~al.(2021)Hendrycks, Burns, Kadavath, Arora, Basart, Tang, Song, and Steinhardt]{math}
Dan Hendrycks, Collin Burns, Saurav Kadavath, Akul Arora, Steven Basart, Eric Tang, Dawn Song, and Jacob Steinhardt.
\newblock Measuring mathematical problem solving with the {MATH} dataset.
\newblock In \emph{Thirty-fifth Conference on Neural Information Processing Systems Datasets and Benchmarks Track (Round 2)}, 2021.
\newblock URL \url{https://openreview.net/forum?id=7Bywt2mQsCe}.

\bibitem[Hu et~al.(2025)Hu, Zhang, Han, Jiang, Zhang, and Shum]{orz}
Jingcheng Hu, Yinmin Zhang, Qi~Han, Daxin Jiang, Xiangyu Zhang, and Heung-Yeung Shum.
\newblock Open-reasoner-zero: An open source approach to scaling up reinforcement learning on the base model.
\newblock \emph{arXiv preprint arXiv:2503.24290}, 2025.

\bibitem[Huan et~al.(2025)Huan, Li, Zheng, Xu, Kim, Du, Poovendran, Neubig, and Yue]{rl-transferability}
Maggie Huan, Yuetai Li, Tuney Zheng, Xiaoyu Xu, Seungone Kim, Minxin Du, Radha Poovendran, Graham Neubig, and Xiang Yue.
\newblock Does math reasoning improve general llm capabilities? understanding transferability of llm reasoning.
\newblock \emph{arXiv preprint arXiv:2507.00432}, 2025.

\bibitem[Jaech et~al.(2024)Jaech, Kalai, Lerer, Richardson, El-Kishky, Low, Helyar, Madry, Beutel, Carney, et~al.]{o1}
Aaron Jaech, Adam Kalai, Adam Lerer, Adam Richardson, Ahmed El-Kishky, Aiden Low, Alec Helyar, Aleksander Madry, Alex Beutel, Alex Carney, et~al.
\newblock Openai o1 system card.
\newblock \emph{arXiv preprint arXiv:2412.16720}, 2024.

\bibitem[Jain et~al.(2025)Jain, Han, Gu, Li, Yan, Zhang, Wang, Solar-Lezama, Sen, and Stoica]{livecodebench}
Naman Jain, King Han, Alex Gu, Wen-Ding Li, Fanjia Yan, Tianjun Zhang, Sida Wang, Armando Solar-Lezama, Koushik Sen, and Ion Stoica.
\newblock Livecodebench: Holistic and contamination free evaluation of large language models for code.
\newblock In \emph{The Thirteenth International Conference on Learning Representations}, 2025.
\newblock URL \url{https://openreview.net/forum?id=chfJJYC3iL}.

\bibitem[Jiang et~al.(2025)Jiang, Xiong, Yuan, Xin, Xu, Yue, Zhao, and Yan]{pag}
Yuhua Jiang, Yuwen Xiong, Yufeng Yuan, Chao Xin, Wenyuan Xu, Yu~Yue, Qianchuan Zhao, and Lin Yan.
\newblock Pag: Multi-turn reinforced llm self-correction with policy as generative verifier.
\newblock \emph{arXiv preprint arXiv:2506.10406}, 2025.

\bibitem[Li et~al.(2023)Li, Zhang, Dubois, Taori, Gulrajani, Guestrin, Liang, and Hashimoto]{alpaca_eval}
Xuechen Li, Tianyi Zhang, Yann Dubois, Rohan Taori, Ishaan Gulrajani, Carlos Guestrin, Percy Liang, and Tatsunori~B. Hashimoto.
\newblock Alpacaeval: An automatic evaluator of instruction-following models.
\newblock \url{https://github.com/tatsu-lab/alpaca_eval}, 5 2023.

\bibitem[Li et~al.(2025)Li, Ma, Li, Li, Rong, Xu, Zhang, Zhao, and Huang]{star-r1}
Zongzhao Li, Zongyang Ma, Mingze Li, Songyou Li, Yu~Rong, Tingyang Xu, Ziqi Zhang, Deli Zhao, and Wenbing Huang.
\newblock Star-r1: Spatial transformation reasoning by reinforcing multimodal llms.
\newblock \emph{arXiv preprint arXiv:2505.15804}, 2025.

\bibitem[Lightman et~al.(2023)Lightman, Kosaraju, Burda, Edwards, Baker, Lee, Leike, Schulman, Sutskever, and Cobbe]{prm800k}
Hunter Lightman, Vineet Kosaraju, Yuri Burda, Harrison Edwards, Bowen Baker, Teddy Lee, Jan Leike, John Schulman, Ilya Sutskever, and Karl Cobbe.
\newblock Let's verify step by step.
\newblock In \emph{The Twelfth International Conference on Learning Representations}, 2023.

\bibitem[Liu et~al.(2024{\natexlab{a}})Liu, Feng, Xue, Wang, Wu, Lu, Zhao, Deng, Zhang, Ruan, et~al.]{deepseek-v3}
Aixin Liu, Bei Feng, Bing Xue, Bingxuan Wang, Bochao Wu, Chengda Lu, Chenggang Zhao, Chengqi Deng, Chenyu Zhang, Chong Ruan, et~al.
\newblock Deepseek-v3 technical report.
\newblock \emph{arXiv preprint arXiv:2412.19437}, 2024{\natexlab{a}}.

\bibitem[Liu et~al.(2024{\natexlab{b}})Liu, Han, Wang, Tsvetkov, Choi, and Smith]{lm_proxy}
Alisa Liu, Xiaochuang Han, Yizhong Wang, Yulia Tsvetkov, Yejin Choi, and Noah~A. Smith.
\newblock Tuning language models by proxy.
\newblock In \emph{First Conference on Language Modeling}, 2024{\natexlab{b}}.
\newblock URL \url{https://openreview.net/forum?id=dribhnhm1i}.

\bibitem[Liu \& Zhang(2025)Liu and Zhang]{code-r1}
Jiawei Liu and Lingming Zhang.
\newblock Code-r1: Reproducing r1 for code with reliable rewards.
\newblock 2025.

\bibitem[Liu et~al.(2025{\natexlab{a}})Liu, Liang, He, Xu, Wang, He, Tu, Mi, and Yu]{trust-but-verify}
Xiaoyuan Liu, Tian Liang, Zhiwei He, Jiahao Xu, Wenxuan Wang, Pinjia He, Zhaopeng Tu, Haitao Mi, and Dong Yu.
\newblock Trust, but verify: A self-verification approach to reinforcement learning with verifiable rewards.
\newblock \emph{arXiv preprint arXiv:2505.13445}, 2025{\natexlab{a}}.

\bibitem[Liu et~al.(2025{\natexlab{b}})Liu, Chen, Li, Qi, Pang, Du, Lee, and Lin]{dr-grpo}
Zichen Liu, Changyu Chen, Wenjun Li, Penghui Qi, Tianyu Pang, Chao Du, Wee~Sun Lee, and Min Lin.
\newblock Understanding r1-zero-like training: A critical perspective.
\newblock \emph{arXiv preprint arXiv:2503.20783}, 2025{\natexlab{b}}.

\bibitem[Lu et~al.(2025)Lu, Wang, Chen, and Tang]{urpo}
Songshuo Lu, Hua Wang, Zhi Chen, and Yaohua Tang.
\newblock Urpo: A unified reward \& policy optimization framework for large language models.
\newblock \emph{arXiv preprint arXiv:2507.17515}, 2025.

\bibitem[Ma et~al.(2025)Ma, Liu, Jiang, Zhang, Ma, and Chen]{general-reasoner}
Xueguang Ma, Qian Liu, Dongfu Jiang, Ge~Zhang, Zejun Ma, and Wenhu Chen.
\newblock General-reasoner: Advancing llm reasoning across all domains.
\newblock \emph{arXiv preprint arXiv:2505.14652}, 2025.

\bibitem[MetaAI(2024{\natexlab{a}})]{llama3.1}
MetaAI.
\newblock Introducing llama 3.1: Our most capable models to date.
\newblock \url{https://ai.meta.com/blog/meta-llama-3-1/}, 2024{\natexlab{a}}.

\bibitem[MetaAI(2024{\natexlab{b}})]{llama3.2}
MetaAI.
\newblock Llama 3.2: Revolutionizing edge ai and vision with open, customizable models.
\newblock \url{https://ai.meta.com/blog/llama-3-2-connect-2024-vision-edge-mobile-devices/}, 2024{\natexlab{b}}.

\bibitem[Mitchell et~al.(2024)Mitchell, Rafailov, Sharma, Finn, and Manning]{emulated_finetuning}
Eric Mitchell, Rafael Rafailov, Archit Sharma, Chelsea Finn, and Christopher~D Manning.
\newblock An emulator for fine-tuning large language models using small language models.
\newblock In \emph{The Twelfth International Conference on Learning Representations}, 2024.
\newblock URL \url{https://openreview.net/forum?id=Eo7kv0sllr}.

\bibitem[Mukherjee et~al.(2025)Mukherjee, Yuan, Hakkani-Tur, and Peng]{rl-finetunes-small-net}
Sagnik Mukherjee, Lifan Yuan, Dilek Hakkani-Tur, and Hao Peng.
\newblock Reinforcement learning finetunes small subnetworks in large language models.
\newblock \emph{arXiv preprint arXiv:2505.11711}, 2025.

\bibitem[OpenCompass(2025)]{aime2025}
OpenCompass.
\newblock Aime 2025.
\newblock \url{https://huggingface.co/datasets/opencompass/AIME2025}, 2025.

\bibitem[Peters \& Schaal(2007)Peters and Schaal]{rl_by_reward_weighted_regression}
Jan Peters and Stefan Schaal.
\newblock Reinforcement learning by reward-weighted regression for operational space control.
\newblock In \emph{Proceedings of the 24th international conference on Machine learning}, pp.\  745--750, 2007.

\bibitem[{Qwen Team}(2024)]{qwen2.5}
{Qwen Team}.
\newblock Qwen2. 5: A party of foundation models.
\newblock \emph{Qwen (Sept. 2024). url: https://qwenlm. github. io/blog/qwen2}, 5, 2024.

\bibitem[Rafailov et~al.(2023)Rafailov, Sharma, Mitchell, Manning, Ermon, and Finn]{dpo}
Rafael Rafailov, Archit Sharma, Eric Mitchell, Christopher~D Manning, Stefano Ermon, and Chelsea Finn.
\newblock Direct preference optimization: Your language model is secretly a reward model.
\newblock \emph{Advances in neural information processing systems}, 36:\penalty0 53728--53741, 2023.

\bibitem[Rein et~al.(2024)Rein, Hou, Stickland, Petty, Pang, Dirani, Michael, and Bowman]{gpqa}
David Rein, Betty~Li Hou, Asa~Cooper Stickland, Jackson Petty, Richard~Yuanzhe Pang, Julien Dirani, Julian Michael, and Samuel~R Bowman.
\newblock Gpqa: A graduate-level google-proof q\&a benchmark.
\newblock In \emph{First Conference on Language Modeling}, 2024.

\bibitem[Sareen et~al.(2025)Sareen, Moss, Sordoni, Agarwal, and Hosseini]{put-value-back-in-rl}
Kusha Sareen, Morgane~M Moss, Alessandro Sordoni, Rishabh Agarwal, and Arian Hosseini.
\newblock Putting the value back in rl: Better test-time scaling by unifying llm reasoners with verifiers.
\newblock \emph{arXiv preprint arXiv:2505.04842}, 2025.

\bibitem[Schulman et~al.(2017)Schulman, Wolski, Dhariwal, Radford, and Klimov]{ppo}
John Schulman, Filip Wolski, Prafulla Dhariwal, Alec Radford, and Oleg Klimov.
\newblock Proximal policy optimization algorithms.
\newblock \emph{arXiv preprint arXiv:1707.06347}, 2017.

\bibitem[Shao et~al.(2024)Shao, Wang, Zhu, Xu, Song, Bi, Zhang, Zhang, Li, Wu, et~al.]{deepseekmath}
Zhihong Shao, Peiyi Wang, Qihao Zhu, Runxin Xu, Junxiao Song, Xiao Bi, Haowei Zhang, Mingchuan Zhang, YK~Li, Yang Wu, et~al.
\newblock Deepseekmath: Pushing the limits of mathematical reasoning in open language models.
\newblock \emph{arXiv preprint arXiv:2402.03300}, 2024.

\bibitem[Sheng et~al.(2024)Sheng, Zhang, Ye, Wu, Zhang, Zhang, Peng, Lin, and Wu]{verl}
Guangming Sheng, Chi Zhang, Zilingfeng Ye, Xibin Wu, Wang Zhang, Ru~Zhang, Yanghua Peng, Haibin Lin, and Chuan Wu.
\newblock Hybridflow: A flexible and efficient rlhf framework.
\newblock \emph{arXiv preprint arXiv: 2409.19256}, 2024.

\bibitem[Skywork-o1(2024)]{skywork-o1}
Skywork-o1.
\newblock Skywork-o1 open series.
\newblock \url{https://huggingface.co/Skywork}, November 2024.
\newblock URL \url{https://huggingface.co/Skywork}.

\bibitem[Snell et~al.(2024)Snell, Lee, Xu, and Kumar]{scaling-test-time-compute}
Charlie Snell, Jaehoon Lee, Kelvin Xu, and Aviral Kumar.
\newblock Scaling llm test-time compute optimally can be more effective than scaling model parameters.
\newblock \emph{arXiv preprint arXiv:2408.03314}, 2024.

\bibitem[Sutton et~al.(1998)Sutton, Barto, et~al.]{rl_introduction}
Richard~S Sutton, Andrew~G Barto, et~al.
\newblock \emph{Reinforcement learning: An introduction}, volume~1.
\newblock MIT press Cambridge, 1998.

\bibitem[Team et~al.(2025{\natexlab{a}})Team, Bai, Bao, Chen, Chen, Chen, Chen, Chen, Chen, Chen, et~al.]{k2}
Kimi Team, Yifan Bai, Yiping Bao, Guanduo Chen, Jiahao Chen, Ningxin Chen, Ruijue Chen, Yanru Chen, Yuankun Chen, Yutian Chen, et~al.
\newblock Kimi k2: Open agentic intelligence.
\newblock \emph{arXiv preprint arXiv:2507.20534}, 2025{\natexlab{a}}.

\bibitem[Team et~al.(2025{\natexlab{b}})Team, Du, Gao, Xing, Jiang, Chen, Li, Xiao, Du, Liao, et~al.]{k1.5}
Kimi Team, Angang Du, Bofei Gao, Bowei Xing, Changjiu Jiang, Cheng Chen, Cheng Li, Chenjun Xiao, Chenzhuang Du, Chonghua Liao, et~al.
\newblock Kimi k1. 5: Scaling reinforcement learning with llms.
\newblock \emph{arXiv preprint arXiv:2501.12599}, 2025{\natexlab{b}}.

\bibitem[Wang et~al.(2024{\natexlab{a}})Wang, Li, Shao, Xu, Dai, Li, Chen, Wu, and Sui]{math-shepherd}
Peiyi Wang, Lei Li, Zhihong Shao, Runxin Xu, Damai Dai, Yifei Li, Deli Chen, Yu~Wu, and Zhifang Sui.
\newblock Math-shepherd: Verify and reinforce llms step-by-step without human annotations.
\newblock In \emph{Proceedings of the 62nd Annual Meeting of the Association for Computational Linguistics (Volume 1: Long Papers)}, pp.\  9426--9439, 2024{\natexlab{a}}.

\bibitem[Wang et~al.(2024{\natexlab{b}})Wang, Ma, Zhang, Ni, Chandra, Guo, Ren, Arulraj, He, Jiang, et~al.]{mmlu-pro}
Yubo Wang, Xueguang Ma, Ge~Zhang, Yuansheng Ni, Abhranil Chandra, Shiguang Guo, Weiming Ren, Aaran Arulraj, Xuan He, Ziyan Jiang, et~al.
\newblock Mmlu-pro: A more robust and challenging multi-task language understanding benchmark.
\newblock \emph{arXiv preprint arXiv:2406.01574}, 2024{\natexlab{b}}.

\bibitem[Wang et~al.(2025{\natexlab{a}})Wang, Zhou, Li, and Liu]{octothinker}
Zengzhi Wang, Fan Zhou, Xuefeng Li, and Pengfei Liu.
\newblock Octothinker: Mid-training incentivizes reinforcement learning scaling.
\newblock \emph{arXiv preprint arXiv:2506.20512}, 2025{\natexlab{a}}.

\bibitem[Wang et~al.(2025{\natexlab{b}})Wang, Wang, Wang, Zhang, Li, Yang, Jin, Yu, Nguyen, Liu, et~al.]{ragen}
Zihan Wang, Kangrui Wang, Qineng Wang, Pingyue Zhang, Linjie Li, Zhengyuan Yang, Xing Jin, Kefan Yu, Minh~Nhat Nguyen, Licheng Liu, et~al.
\newblock Ragen: Understanding self-evolution in llm agents via multi-turn reinforcement learning.
\newblock \emph{arXiv preprint arXiv:2504.20073}, 2025{\natexlab{b}}.

\bibitem[Wen et~al.(2025)Wen, Liu, Zheng, Xu, Ye, Wu, Liang, Wang, Li, Miao, et~al.]{rlvr-correct-cot}
Xumeng Wen, Zihan Liu, Shun Zheng, Zhijian Xu, Shengyu Ye, Zhirong Wu, Xiao Liang, Yang Wang, Junjie Li, Ziming Miao, et~al.
\newblock Reinforcement learning with verifiable rewards implicitly incentivizes correct reasoning in base llms.
\newblock \emph{arXiv preprint arXiv:2506.14245}, 2025.

\bibitem[Yang et~al.(2024)Yang, Zhang, Hui, Gao, Yu, Li, Liu, Tu, Zhou, Lin, Lu, Xue, Lin, Liu, Ren, and Zhang]{qwen2.5-math}
An~Yang, Beichen Zhang, Binyuan Hui, Bofei Gao, Bowen Yu, Chengpeng Li, Dayiheng Liu, Jianhong Tu, Jingren Zhou, Junyang Lin, Keming Lu, Mingfeng Xue, Runji Lin, Tianyu Liu, Xingzhang Ren, and Zhenru Zhang.
\newblock Qwen2.5-math technical report: Toward mathematical expert model via self-improvement.
\newblock \emph{arXiv preprint arXiv:2409.12122}, 2024.

\bibitem[Yang et~al.(2025{\natexlab{a}})Yang, Li, Yang, Zhang, Hui, Zheng, Yu, Gao, Huang, Lv, et~al.]{qwen3}
An~Yang, Anfeng Li, Baosong Yang, Beichen Zhang, Binyuan Hui, Bo~Zheng, Bowen Yu, Chang Gao, Chengen Huang, Chenxu Lv, et~al.
\newblock Qwen3 technical report.
\newblock \emph{arXiv preprint arXiv:2505.09388}, 2025{\natexlab{a}}.

\bibitem[Yang et~al.(2025{\natexlab{b}})Yang, Chen, Lin, and Wen]{deepcritic}
Wenkai Yang, Jingwen Chen, Yankai Lin, and Ji-Rong Wen.
\newblock Deepcritic: Deliberate critique with large language models.
\newblock \emph{arXiv preprint arXiv:2505.00662}, 2025{\natexlab{b}}.

\bibitem[Yu et~al.(2025{\natexlab{a}})Yu, Zhang, Zhu, Yuan, Zuo, Yue, Dai, Fan, Liu, Liu, et~al.]{dapo}
Qiying Yu, Zheng Zhang, Ruofei Zhu, Yufeng Yuan, Xiaochen Zuo, Yu~Yue, Weinan Dai, Tiantian Fan, Gaohong Liu, Lingjun Liu, et~al.
\newblock Dapo: An open-source llm reinforcement learning system at scale.
\newblock \emph{arXiv preprint arXiv:2503.14476}, 2025{\natexlab{a}}.

\bibitem[Yu et~al.(2025{\natexlab{b}})Yu, Ji, Wang, Yao, Wang, Cui, Yuan, Ding, Yao, Liu, et~al.]{rlpr}
Tianyu Yu, Bo~Ji, Shouli Wang, Shu Yao, Zefan Wang, Ganqu Cui, Lifan Yuan, Ning Ding, Yuan Yao, Zhiyuan Liu, et~al.
\newblock Rlpr: Extrapolating rlvr to general domains without verifiers.
\newblock \emph{arXiv preprint arXiv:2506.18254}, 2025{\natexlab{b}}.

\bibitem[Yuan et~al.(2024)Yuan, Li, Chen, Cui, Ding, Zhang, Zhou, Liu, and Peng]{implicitprm}
Lifan Yuan, Wendi Li, Huayu Chen, Ganqu Cui, Ning Ding, Kaiyan Zhang, Bowen Zhou, Zhiyuan Liu, and Hao Peng.
\newblock Free process rewards without process labels.
\newblock \emph{arXiv preprint arXiv:2412.01981}, 2024.

\bibitem[Yue et~al.(2025{\natexlab{a}})Yue, Chen, Lu, Zhao, Wang, Song, and Huang]{passk}
Yang Yue, Zhiqi Chen, Rui Lu, Andrew Zhao, Zhaokai Wang, Shiji Song, and Gao Huang.
\newblock Does reinforcement learning really incentivize reasoning capacity in llms beyond the base model?
\newblock \emph{arXiv preprint arXiv:2504.13837}, 2025{\natexlab{a}}.

\bibitem[Yue et~al.(2025{\natexlab{b}})Yue, Yuan, Yu, Zuo, Zhu, Xu, Chen, Wang, Fan, Du, et~al.]{vapo}
Yu~Yue, Yufeng Yuan, Qiying Yu, Xiaochen Zuo, Ruofei Zhu, Wenyuan Xu, Jiaze Chen, Chengyi Wang, TianTian Fan, Zhengyin Du, et~al.
\newblock Vapo: Efficient and reliable reinforcement learning for advanced reasoning tasks.
\newblock \emph{arXiv preprint arXiv:2504.05118}, 2025{\natexlab{b}}.

\bibitem[Zha et~al.(2025)Zha, Gao, Shen, Hong, Boning, and Katabi]{rl-tango}
Kaiwen Zha, Zhengqi Gao, Maohao Shen, Zhang-Wei Hong, Duane~S Boning, and Dina Katabi.
\newblock Rl tango: Reinforcing generator and verifier together for language reasoning.
\newblock \emph{arXiv preprint arXiv:2505.15034}, 2025.

\bibitem[Zhang et~al.(2024)Zhang, Hosseini, Bansal, Kazemi, Kumar, and Agarwal]{generative_verifiers}
Lunjun Zhang, Arian Hosseini, Hritik Bansal, Mehran Kazemi, Aviral Kumar, and Rishabh Agarwal.
\newblock Generative verifiers: Reward modeling as next-token prediction.
\newblock \emph{arXiv preprint arXiv:2408.15240}, 2024.

\bibitem[Zhang et~al.(2025)Zhang, Zheng, Wu, Zhang, Lin, Yu, Liu, Zhou, and Lin]{lessons_of_prm}
Zhenru Zhang, Chujie Zheng, Yangzhen Wu, Beichen Zhang, Runji Lin, Bowen Yu, Dayiheng Liu, Jingren Zhou, and Junyang Lin.
\newblock The lessons of developing process reward models in mathematical reasoning.
\newblock \emph{arXiv preprint arXiv:2501.07301}, 2025.

\bibitem[Zhao et~al.(2025)Zhao, Liu, Zhang, Zhou, Gao, Li, Lyu, Qian, Qi, Li, et~al.]{genprm}
Jian Zhao, Runze Liu, Kaiyan Zhang, Zhimu Zhou, Junqi Gao, Dong Li, Jiafei Lyu, Zhouyi Qian, Biqing Qi, Xiu Li, et~al.
\newblock Genprm: Scaling test-time compute of process reward models via generative reasoning.
\newblock \emph{arXiv preprint arXiv:2504.00891}, 2025.

\bibitem[Zheng et~al.(2025)Zheng, Liu, Li, Chen, Yu, Gao, Dang, Liu, Men, Yang, et~al.]{gspo}
Chujie Zheng, Shixuan Liu, Mingze Li, Xiong-Hui Chen, Bowen Yu, Chang Gao, Kai Dang, Yuqiong Liu, Rui Men, An~Yang, et~al.
\newblock Group sequence policy optimization.
\newblock \emph{arXiv preprint arXiv:2507.18071}, 2025.

\bibitem[Zhou et~al.(2025)Zhou, Liu, Sims, Wang, Pang, Li, Wang, Lin, and Du]{verifree}
Xiangxin Zhou, Zichen Liu, Anya Sims, Haonan Wang, Tianyu Pang, Chongxuan Li, Liang Wang, Min Lin, and Chao Du.
\newblock Reinforcing general reasoning without verifiers.
\newblock \emph{arXiv preprint arXiv:2505.21493}, 2025.

\bibitem[Zuo et~al.(2025)Zuo, Zhang, Sheng, Qu, Cui, Zhu, Li, Zhang, Long, Hua, et~al.]{TTRL}
Yuxin Zuo, Kaiyan Zhang, Li~Sheng, Shang Qu, Ganqu Cui, Xuekai Zhu, Haozhan Li, Yuchen Zhang, Xinwei Long, Ermo Hua, et~al.
\newblock Ttrl: Test-time reinforcement learning.
\newblock \emph{arXiv preprint arXiv:2504.16084}, 2025.

\end{thebibliography}
\bibliographystyle{iclr2025_conference}

\clearpage
\appendix

\section{The Length Bias in Implicit Reward}
\label{appendix: length bias of implicit reward}

\begin{figure}[t]  
\begin{center}
\begin{minipage}[t]{0.49\linewidth}
\centerline{\includegraphics[width=1\linewidth]{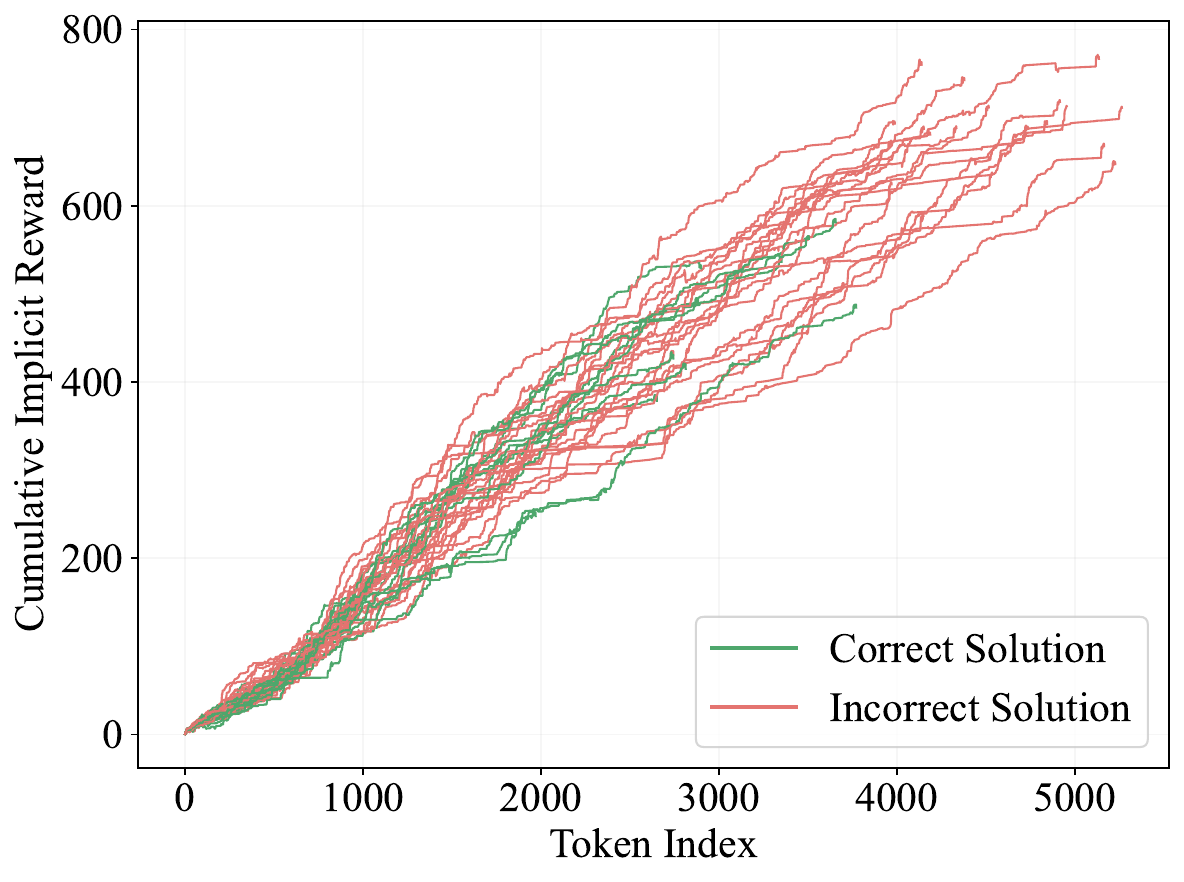}}
\caption{Cumulative implicit reward values across 32 reasoning trajectories sampled from Open-Reasoner-Zero-7B on an AIME2024 problem. Red lines correspond to wrong solutions and green lines correspond to correct solutions.}
\label{fig: implicit reward}
\end{minipage}  
\hfill
\begin{minipage}[t]{0.49\linewidth}
\centerline{\includegraphics[width=1\linewidth]{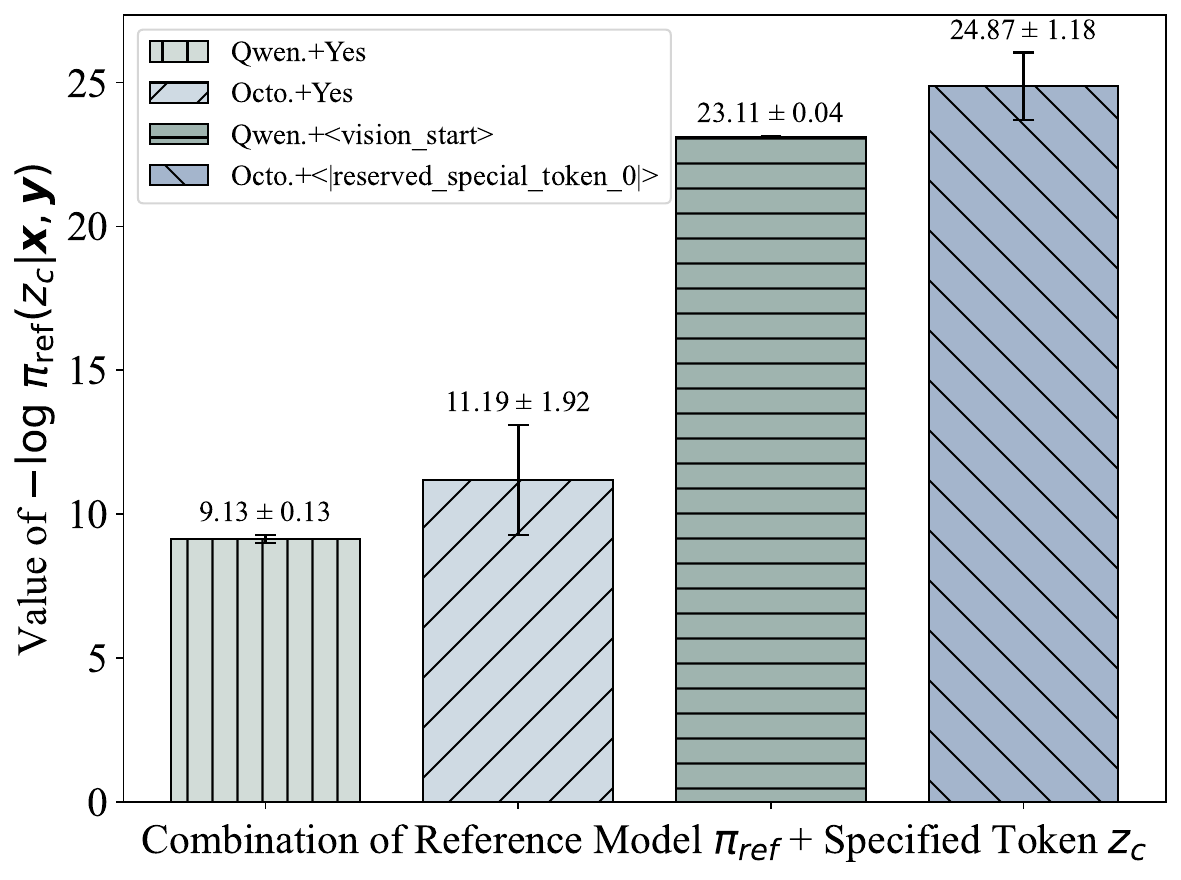}}
\caption{The mean and standard deviation of $-\log \pi_{ref}(z_{c}|\boldsymbol{x},\boldsymbol{y})$ under different combinations of reference model $\pi_{ref}$ and pre-specified token $z_{c}$ over 300 input-output pairs.}
\label{fig: small value of log ref}
\end{minipage}  
\end{center}
\end{figure}

Here, we present the trend of the cumulative implicit reward values ($\log \frac{\pi_{\boldsymbol{\theta}}(\boldsymbol{y}_{<i}|\boldsymbol{x})}{\pi_{ref}(\boldsymbol{y}_{<i}|\boldsymbol{x})}$ where $\pi_{ref}$ is Qwen2.5-7B-Base) across 32 reasoning trajectories sampled from Open-Reasoner-Zero-7B on an AIME2024 problem, showing how they vary with the increasing trajectory lengths. As illustrated in Figure~\ref{fig: implicit reward}, the curves of all samples exhibit a positive correlation between the implicit reward and the number of tokens, and longer trajectories tend to yield higher final implicit reward scores, indicating a strong length bias in implicit reward. Since incorrect solutions are generally longer than correct ones in reasoning tasks~\citep{dont-overthink}, implicit reward is therefore not a reliable indicator of the relative quality of reasoning paths at test time.

\section{Statistics of \texorpdfstring{$\log \pi_{ref}(z_{c}|\boldsymbol{x},\boldsymbol{y})$}.}
\label{appendix: small value of log ref}

We present the mean and standard deviation of $-\log \pi_{ref}(z_{c} | \boldsymbol{x},\boldsymbol{y})$ computed over 300 input-output pairs. The reference model $\pi_{ref}$ is chosen as either Qwen2.5-7B-Base or OctoThinker-3B-Short-Base, and the evaluation is performed under two different choices of $z_{c}$ for each reference model (one common token and one unused special token): ``\texttt{Yes}'' and ``\texttt{<vision\_start>}'' for Qwen2.5-7B-Base, ``\texttt{Yes}'' and ``\texttt{<|reserved\_special\_token\_0|>}'' for OctoThinker-3B-Short-Base. The results in Figure~\ref{fig: small value of log ref} indicates that $-\log \pi_{ref}(z_{c} | \boldsymbol{x},\boldsymbol{y})$ remains nearly constant and extremely small, with only a low standard deviation across different $\boldsymbol{x}$ and $\boldsymbol{y}$. Thus, we can consider $\log \pi_{ref}(z_{c} | \boldsymbol{x},\boldsymbol{y})$ as a constant when calculating the last-token self-rewarding scores, which effectively reduces the computational cost by half.

\section{Detailed Training Settings}
\label{appendix: training settings}

\begin{table}[t]  
\begin{center}
\begin{minipage}[t]{0.48\linewidth}
\caption{Basic training hyper-parameters of both GRPO and LaSeR.}
\label{tab: basic hyper-parameters}
\centering
\begin{tabular}{ll}
\toprule
Hyper-parameter & Value \\
\midrule
Train Batch Size & 128  \\
Micro Batch Size & 128 \\
Rollout $n$ & 8 \\
Maximum Prompt Length & 2048 \\
Maximum Response Length & 8192 \\
Temperature & 1.0 \\
Top $p$ & 1.0 \\
LR & $1\times 10^{-6}$ \\
KL Coefficient & 0.0 \\
\bottomrule
\end{tabular}
\end{minipage}  
\hfil
\begin{minipage}[t]{0.48\linewidth}
\caption{Unique training hyper-parameters of LaSeR.}
\label{tab: unique hyper-parameters}
\centering
\begin{tabular}{ll}
\toprule
Hyper-parameter & Value \\
\midrule
Coefficient $\beta_v$ & 0.1  \\
Loss Weight $\alpha$ & 0.1 \\
Self-Rewarding Adv. Weight $\tau$ & 0.1\\
Reasoning Warm-Up Steps & 200 \\
Self-Rewarding Warm-Up Steps & 200\\
\bottomrule
\end{tabular}
\end{minipage} 
\end{center}
\end{table}
 
We use \texttt{verl}~\citep{verl} as our RL training framework. The basic training hyper-parameters in both GRPO training and LaSeR training for each model are put in Table~\ref{tab: basic hyper-parameters}, and the newly introduced training hyper-parameters for LaSeR are put in Table~\ref{tab: unique hyper-parameters}. The number of optimization steps is 1000 for Qwen2.5-7B-Base and OctoThinker-3B-Short-Base, and 500 for Open-Reasoner-Zero-7B.  
In RL, a reasoning reward of 1.0 is given if the final answer and the answer format are both correct; otherwise, it is 0.0. In our method, the reasoning warm-up is performed for Qwen2.5-7B-Base and OctoThinker-3B-Short-Base only, and the self-rewarding warm-up is performed for all models.

\section{Ablation Studies on Self-Rewarding Hyper-Parameters}
\label{appendix: ablation studies on self-rewarding hyper-parameters}

\begin{figure*}[t]
  \centering
  \begin{tabular}{c}
\includegraphics[width=0.95\linewidth]{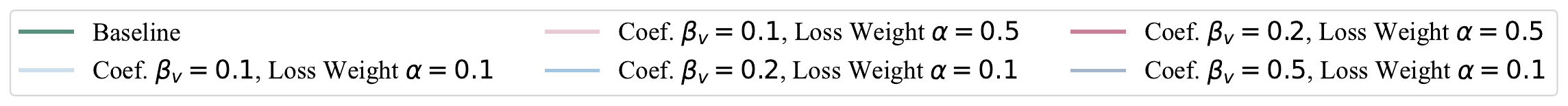}
\end{tabular}
\vskip -0.1in
  \subfigure[Training rewards with EMA smoothing]{\includegraphics[width=0.98\textwidth]{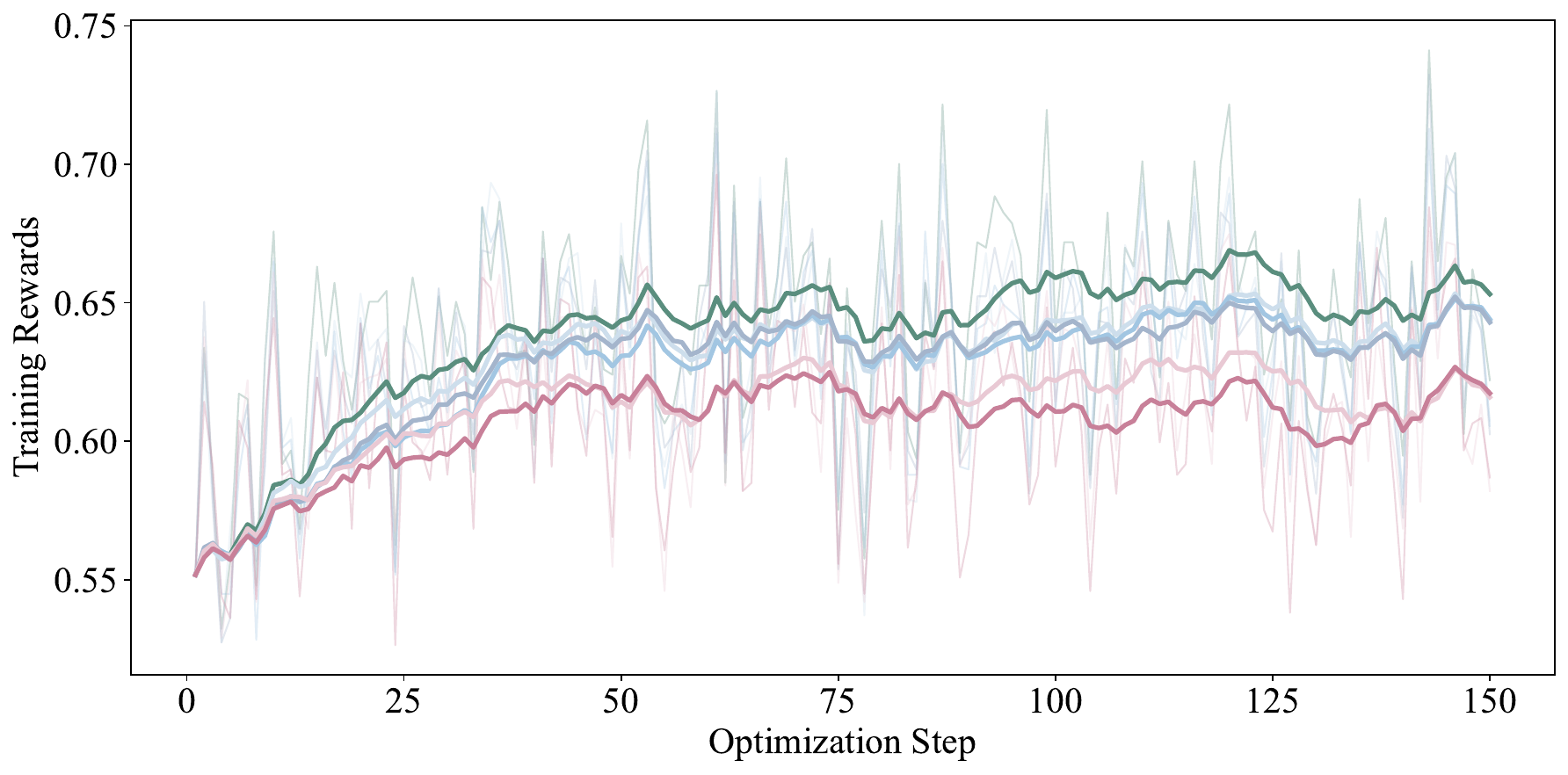}
  }
  \subfigure[Training self-verification F1 scores with EMA smoothing]{\includegraphics[width=0.98\textwidth]{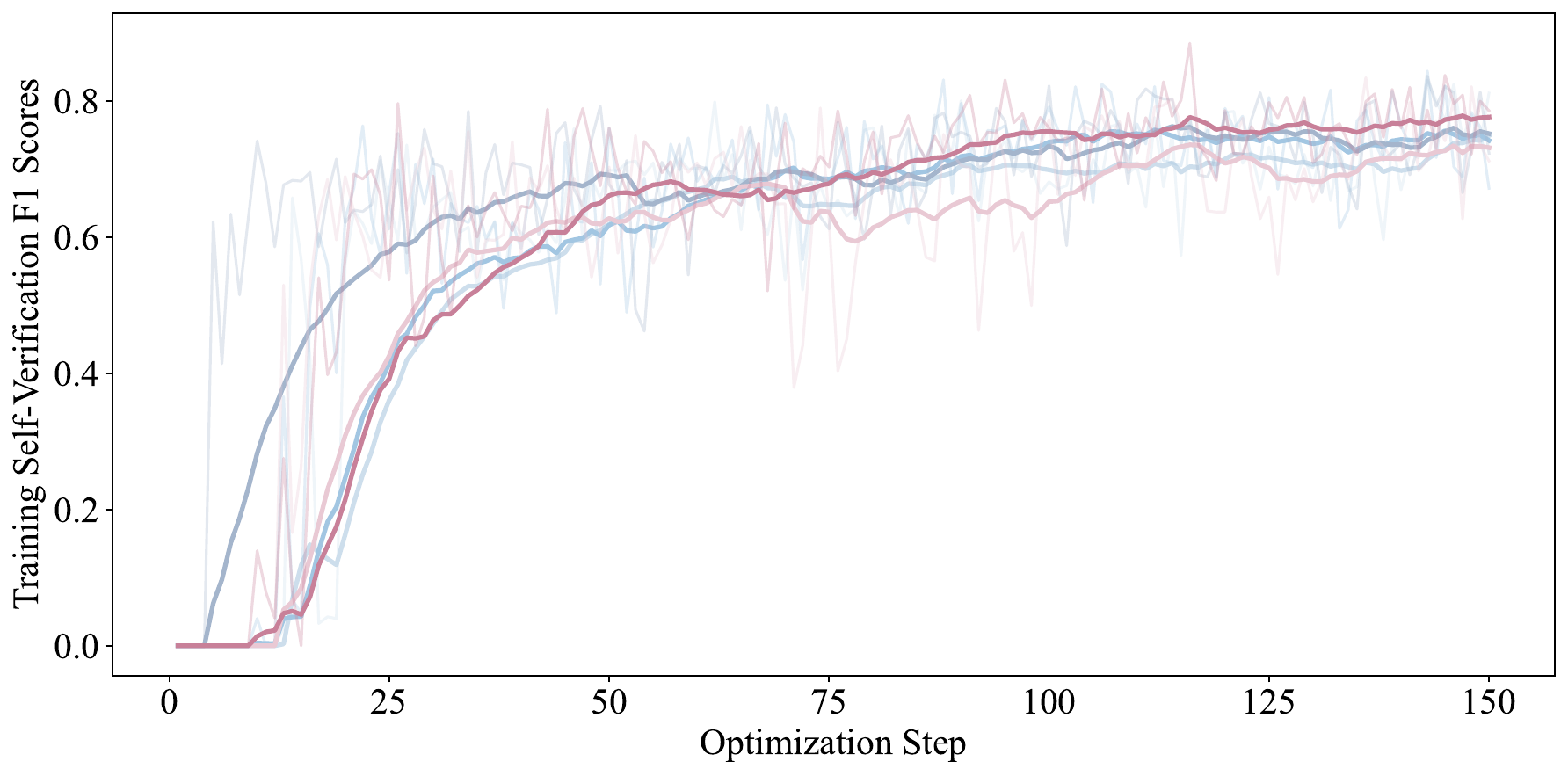}
  }
  \caption{The curves of training rewards and training self-verification F1 scores under different combinations of hyper-parameters with EMA smoothing (EMA coef.=0.9).
  }
  \label{fig: ablation on hyper-parameters}
\end{figure*}

Here, we display the curves (with Exponential Moving Average (EMA) smoothing) of training rewards and training self-verification F1 scores of our method under different choices of coefficient $\beta_v$ and self-rewarding MSE loss weight $\alpha$. The experiments are conducted on Open-Reasoner-Zero-7B, which help to skip the reasoning warm-up phase compared with using Qwen2.5-7B-Base and OctoThinker-3B-Short-Base, while the results are similar in other two base models after reasoning warm-up. The dynamics of training rewards and training self-verification F1 scores are displayed in Figure~\ref{fig: ablation on hyper-parameters}. As we can see, assigning a larger weight $\alpha$ to the last-token self-rewarding loss has a more detrimental impact on the model’s reasoning capabilities. On the other hand, the coefficient $\beta_{v}$ has little impact on optimizing the self-rewarding scores, as long as it remains within a reasonable range ($0.1 \sim 0.5$). However, much smaller values of $\beta_{v}$ can impair the model’s reasoning capability, as indicated by the analysis in the end of Section~\ref{subsec: brief discussion}. For example, when $\beta_v=0.05$, we should have $\pi_{\boldsymbol{\theta}} (z_{c} | \boldsymbol{x}, \boldsymbol{y})=e^{-3}\approx 0.05$ under
\(\pi_{\text{ref}} (z_{c} | \boldsymbol{x}, \boldsymbol{y}) = e^{-23}\) and $r_{v}(\boldsymbol{x},\boldsymbol{y})=1$, then the large value of $\pi_{\boldsymbol{\theta}} (z_{c} | \boldsymbol{x}, \boldsymbol{y})$ causes large
interference with the optimization of reasoning capability. In our main experiments, we choose $(\beta_{v},\alpha)=(0.1,0.1)$.

\section{The Effect of Class-Level Re-Weighting on The Balanced Self-Verification Capability}
\label{appendix: effect of class-level re-weighting}

\begin{figure*}[t]
  \centering
  \begin{tabular}{c}
\includegraphics[width=0.95\linewidth]{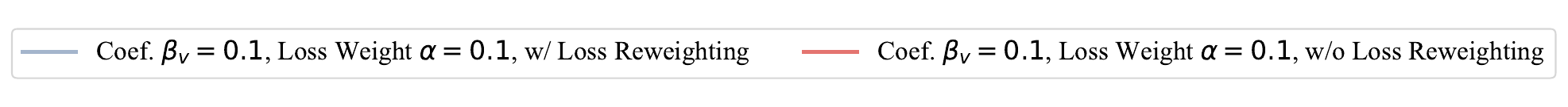}
\end{tabular}
\vskip -0.1in
  \subfigure[Training rewards with EMA smoothing]{\includegraphics[width=0.98\textwidth]{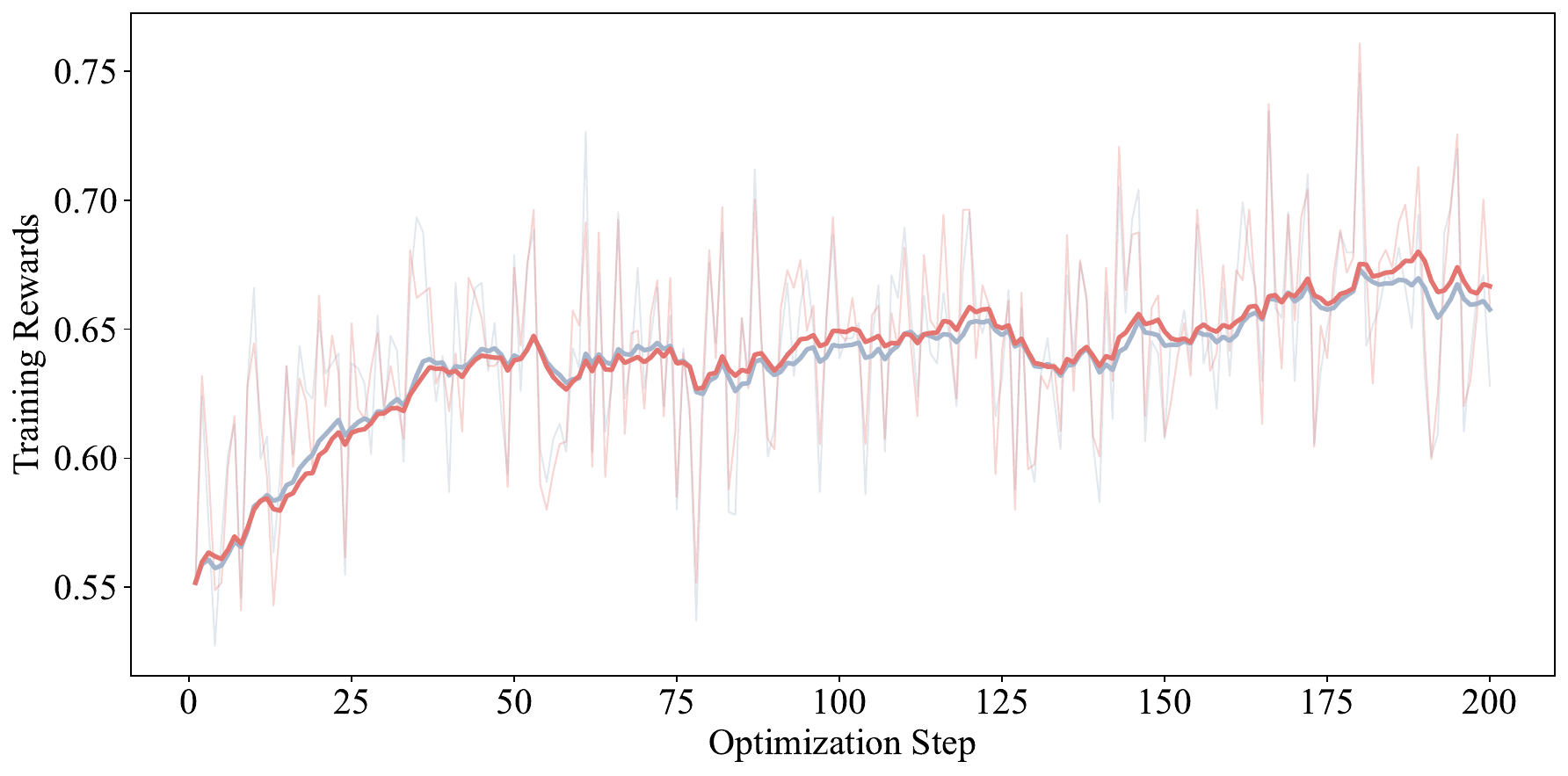}
  }
  \subfigure[Training self-verification F1 scores with EMA smoothing]{\includegraphics[width=0.98\textwidth]{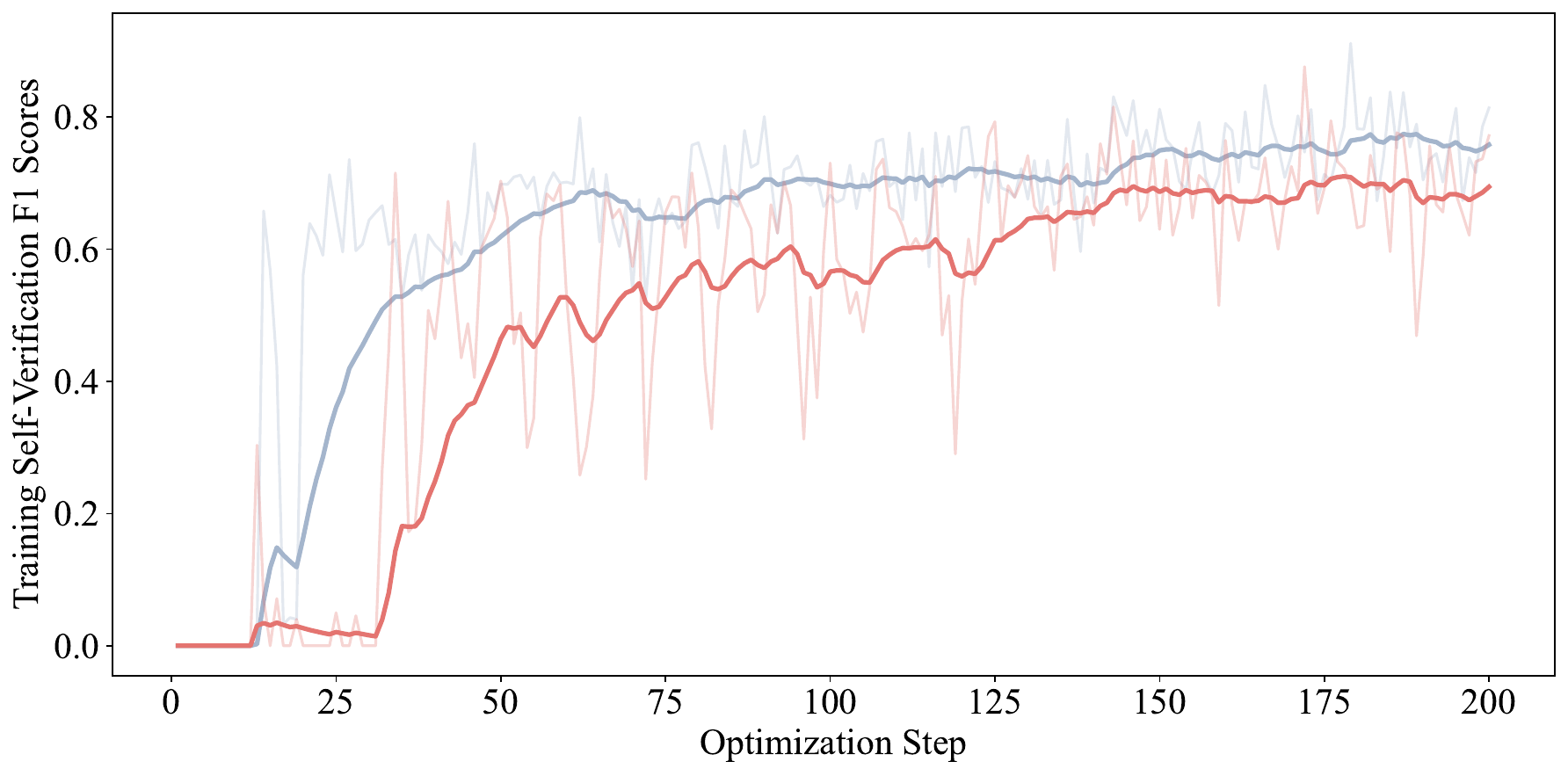}
  }
  \caption{The curves of training rewards and training self-verification F1 scores of our method with and without class-level loss re-weighting practice (EMA coef.=0.9).
  }
  \label{fig: ablation on reweighting}
\end{figure*}

We present the training dynamics of our method on Open-Reasoner-Zero-7B, with and without class-level loss re-weighting, in Figure~\ref{fig: ablation on reweighting} for comparison. As shown, applying loss re-weighting leads to a more balanced self-verification performance by mitigating the bias toward the majority class with larger sample size, while still maintaining high reasoning accuracy.

\section{Comparison between Last-Token Self-Rewarding Loss and Supervised Fine-Tuning Loss}
\label{appendix: comparison with SFT loss}
\begin{figure*}[t]
  \centering
  \begin{tabular}{c}
\includegraphics[width=0.95\linewidth]{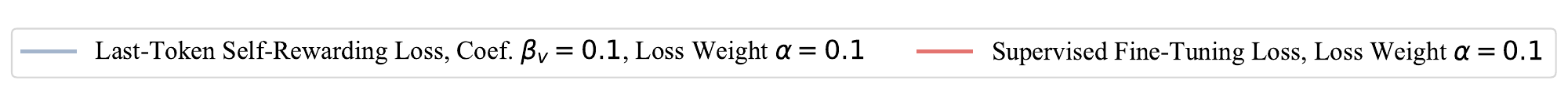}
\end{tabular}
\vskip -0.1in
  \subfigure[Training rewards with EMA smoothing]{\includegraphics[width=0.98\textwidth]{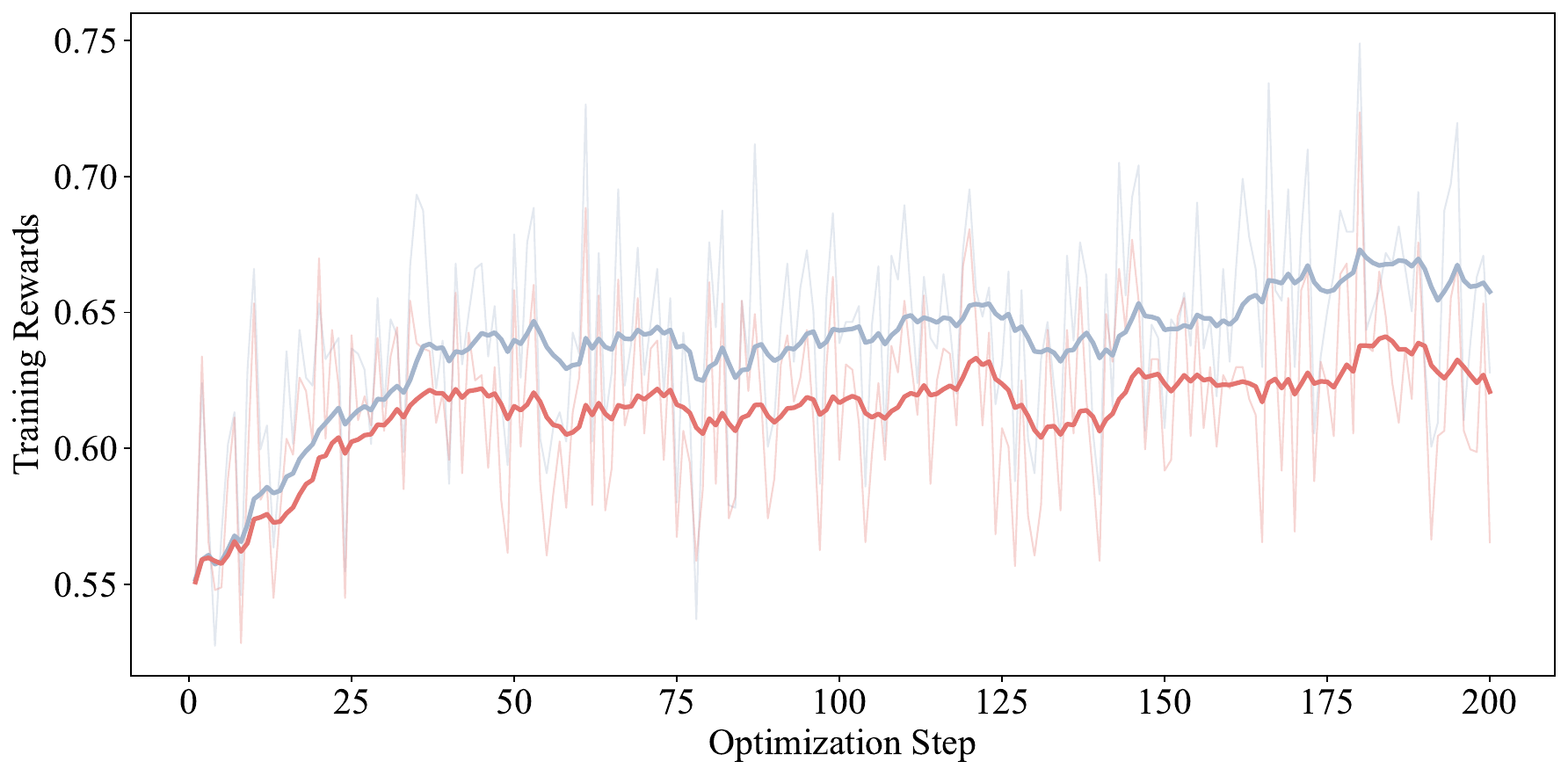}
  }
  \subfigure[Self-rewarding/SFT loss curve on a log scale]{\includegraphics[width=0.98\textwidth]{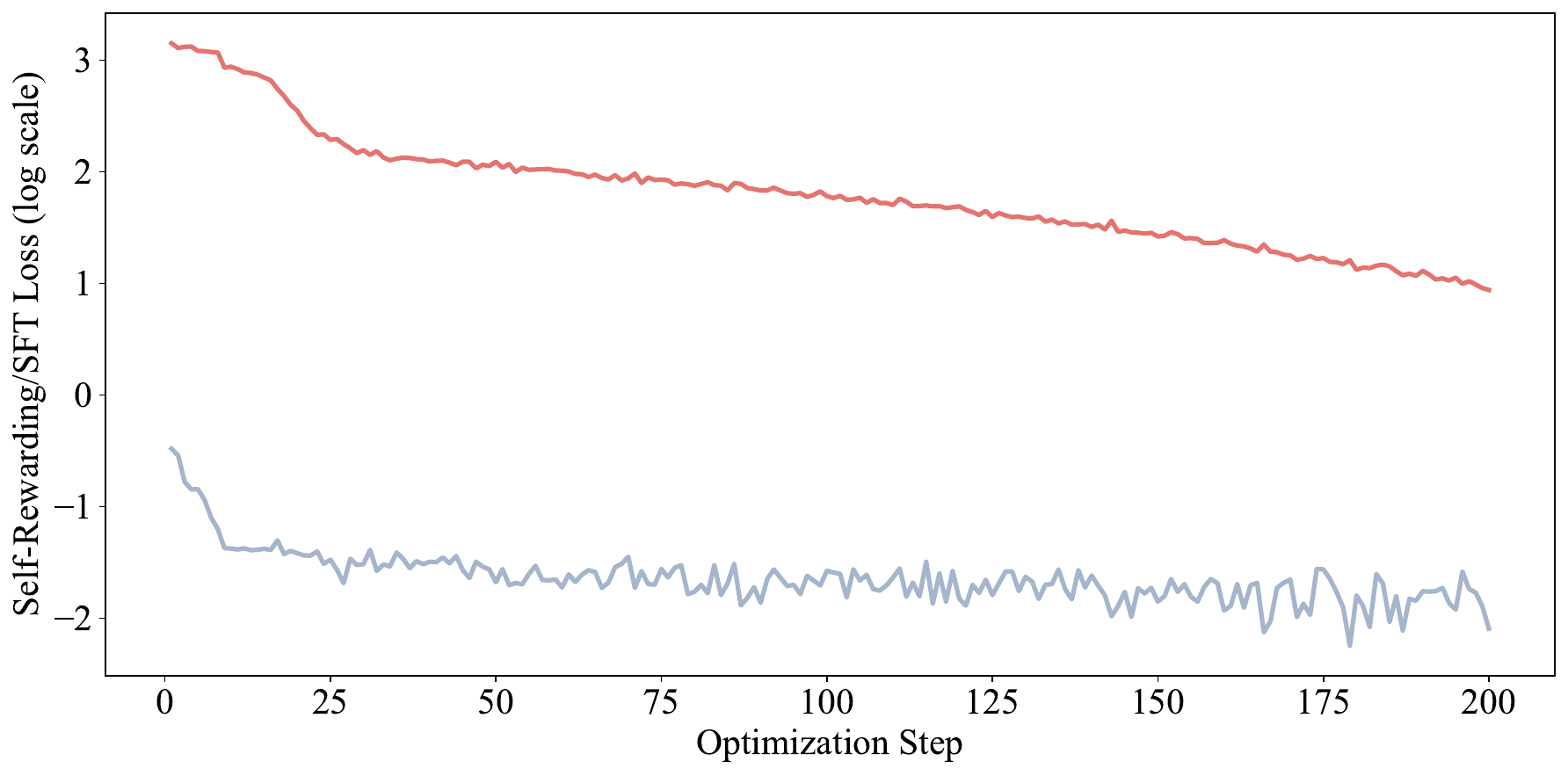}
  }
  \caption{The comparison of the training dynamics between the last-token self-rewarding loss and the SFT loss.
  }
  \label{fig: comparison with sft loss}
\end{figure*}

Following the discussion in Section~\ref{subsec: brief discussion}, we compare the training performance of our introduced last-token self-rewarding loss with the supervised fine-tuning (SFT) loss on Open-Reasoner-Zero-7B. The training dynamics are shown in Figure~\ref{fig: comparison with sft loss}. As observed, applying the SFT loss to optimize the self-rewarding capability causes substantial interference with the optimization of reasoning capability, leading to a marked degradation in training rewards. Moreover, the SFT loss degrades extremely slowly, indicating that directly driving $\pi_{\boldsymbol{\theta}} (z_{c}|\boldsymbol{x},\boldsymbol{y})$ from $0$ to $1$ for \(r_{v}(\boldsymbol{x},\boldsymbol{y}) = 1\) is inherently difficult. However, our method only requires fitting $\pi_{\boldsymbol{\theta}}(z_{c}|\boldsymbol{x},\boldsymbol{y})$ to \(\exp(1 / \beta_{v}) \cdot \pi_{\text{ref}} (z_{c} | \boldsymbol{x}, \boldsymbol{y})\) for \(r_{v}(\boldsymbol{x},\boldsymbol{y}) = 1\), which is considerably easier and introduces much less interference.

\section{Detailed Self-Verification Results}
\label{appendix: self-verification results}
\begin{table*}[t]
\caption{Detailed self-verification results.}
\label{tab: self-verification results}
\centering
\begin{tabular}{lcccccccccc}
\toprule
\multirow{2.5}{*}{\begin{tabular}[c]{@{}l@{}}Method \end{tabular}} & \multicolumn{2}{c}{MATH500} & \multicolumn{2}{c}{AMC23} & \multicolumn{2}{c}{AIME24} & \multicolumn{2}{c}{AIME25} & \multicolumn{2}{c}{Olym.}  \\
\cmidrule(lr){2-3}
\cmidrule(lr){4-5}
\cmidrule(lr){6-7}
\cmidrule(lr){8-9}
\cmidrule(lr){10-11}
&  Acc. &  F1 &  Acc. &  F1 &  Acc. &  F1 &  Acc. &  F1 &  Acc. &  F1 \\
\midrule
\multicolumn{11}{l}{\emph{\quad \textbf{OctoThinker-3B-Short-Base}}} \\
Base & 60.2 & 22.3 & 52.3 & 11.2 & - &  - & - & - & 62.0 & 13.7 \\
GRPO & 58.2 & 56.9 & 66.7 & 47.3 & - & - & - & - & 66.4 & 48.8 \\
\rowcolor{cyan!10}  LaSeR  & 77.0 & 73.6 & 77.3 & 70.2 &  -& -  & - & - & 80.3 &  73.6 \\
\rowcolor{cyan!10} ~~ \scriptsize{- \textit{SWA}} & 81.0 & 80.4 & 84.1 & 70.9 & - & - &- & -& 83.5 & 66.0 \\
\midrule
\multicolumn{11}{l}{\emph{\quad \textbf{Qwen2.5-7B-Base}}} \\
Base & 45.0 &  36.4 & 30.7 & 30.8 & 24.5 & 27.6  & 28.2 & 32.9 & 33.8 & 36.9 \\
GRPO &  76.5 & 54.6 & 61.1 & 59.7 & 60.4 & 36.6 & 72.5 & 41.5 & 54.6 & 53.5 \\
\rowcolor{cyan!10}   LaSeR & 88.0 & 83.2 & 81.5 & 82.5 & 92.2 & 79.6 & 90.5 & 74.3 & 79.5 & 78.3   \\
\rowcolor{cyan!10} ~~ \scriptsize{- \textit{SWA}} & 87.8 & 79.7 &  79.6 & 80.2 & 94.3 & 81.3 & 92.2 & 74.9 & 83.9 & 83.3 \\
\midrule
\multicolumn{11}{l}{\emph{\quad \textbf{Open-Reasoner-Zero-7B}}} \\
Base & 79.6 & 26.7 & 66.6  &  51.3 & 39.6  & 45.9 & 47.6 &  55.2 & 55.2 & 37.5 \\
GRPO & 52.9 & 57.1 & 50.9  & 44.8 & 66.9 & 14.6 & 78.9 & 28.1   & 54.7  & 49.5 \\
\rowcolor{cyan!10}   LaSeR & 90.1 & 87.2 & 77.7 & 79.7 & 87.2 & 64.6 & 92.8 & 77.7 & 80.1 & 78.7 \\
\rowcolor{cyan!10} ~~ \scriptsize{- \textit{SWA}} & 89.0 & 87.5 &  76.2 & 77.7 & 87.7 & 63.3 & 93.6 & 77.3 & 80.2 & 77.9 \\
\bottomrule
\end{tabular}
\end{table*}

We report the detailed self-verification results of each model on self-generated solutions across all benchmarks in Table~\ref{tab: self-verification results}, including both overall accuracy and F1 score. Our method consistently yields significant improvements in model's self-rewarding and self-verification capabilities, while incurring only minimal additional computational cost.~\looseness=-1

\section{Training and Evaluation Settings in General Reasoning Experiments}
\label{appendix: settings on general reasoning tasks}
The basic training and testing hyper-parameters for experiments on WebInstruct-verified are the same as those in Table~\ref{tab: basic hyper-parameters} and Table~\ref{tab: unique hyper-parameters}, while the number of optimization steps here is 800. The simplified constant of the reference log-probability $c_{\text{ref}}$ is $-23.0$.  We do not employ the advantage integration strategy here, as we find that the optimized self-rewarding capability of Qwen3-4B-LaSeR on general reasoning tasks is limited, and introducing self-rewarding-based advantage integration leads to performance degradation.

\section{Prompt Templates}
\label{appendix: prompt templates}
We show the training, evaluation and self-verification prompt templates used in our experiments in the end.
\clearpage
\begin{samepage}
\begin{prompt}[title={Training and Evaluation Prompt Template for OctoThinker-3B-Short-Base}]
\label{prompt: octothinker}
  $<$bos\_token$>$ A conversation between User and Assistant. The user asks a question, and the Assistant solves it. The assistant first thinks about the reasoning process in the mind and then provides the user with the answer. 
  \\
  User: You must put your answer inside \verb|\|boxed\{\} and Your final answer will be extracted automatically by the \verb|\|boxed\{\}  tag.
  \\
  \{question\}
  \\
  Assistant:
\end{prompt}
\end{samepage}
\begin{samepage}
\begin{prompt}[title={Training Prompt Template for Qwen2.5-7B-Base}]
\label{prompt: qwen2.5}
  $<$bos\_token$>$ A conversation between User and Assistant. The User asks a question, and the Assistant solves it. The Assistant first thinks about the reasoning process in the mind and then provides the User with the answer. The reasoning process is enclosed within $<$think$>$ $<$/think$>$ and answer is enclosed within $<$answer$>$ $<$/answer$>$ tags, respectively, i.e., $<$think$>$ reasoning process here $<$/think$>$ $<$answer$>$ answer here $<$/answer$>$.
  \\
  User: You must put your answer inside $<$answer$>$ $<$/answer$>$ tags, i.e., $<$answer$>$ answer here $<$/answer$>$. And your final answer will be extracted automatically by the \verb|\|boxed\{\} tag.
  \\
  This is the problem:
  \\
 \{question\}
 \\
  Assistant: $<$think$>$
\end{prompt}
\end{samepage}
\begin{samepage}
\begin{prompt}[title={Zero-Shot Evaluation Prompt Template for Qwen2.5-7B-Base}]
\label{prompt: qwen2.5 eval}
$<|$im\_start$|>$system
\\
You are a helpful assistant.$<|$im\_end$|>$
\\
$<|$im\_start$|>$user
\\
\{question\}
\\
Please reason step by step, and put your final answer within  \verb|\|boxed\{\}.$<|$im\_end$|>$
\\
$<|$im\_start$|>$assistant
\end{prompt}
\end{samepage}
\begin{samepage}
\begin{prompt}[title={Training and Evaluation Prompt Template for Open-Reasoner-Zero-7B}]
\label{prompt: orz}
  A conversation between User and Assistant. The User asks a question, and the Assistant solves it. The Assistant first thinks about the reasoning process in the mind and then provides the User with the answer. The reasoning process is enclosed within $<$think$>$ $<$/think$>$ and answer is enclosed within $<$answer$>$ $<$/answer$>$ tags, respectively, i.e., $<$think$>$ reasoning process here $<$/think$>$ $<$answer$>$ answer here $<$/answer$>$.
  \\
  User: You must put your answer inside $<$answer$>$ $<$/answer$>$ tags, i.e., $<$answer$>$ answer here $<$/answer$>$. And your final answer will be extracted automatically by the \verb|\|boxed\{\} tag.
  \\
 \{question\}
 \\
  Assistant: $<$think$>$
\end{prompt}
\end{samepage}
\begin{samepage}
\begin{prompt}[title={Training and Evaluation Prompt Template for Qwen3-4B-Base}]
\label{prompt: qwen3}
$<|$im\_start$|>$user
\\
\{question\}
\\
Please reason step by step, and put your final answer within  \verb|\|boxed\{\}.$<|$im\_end$|>$
\\
$<|$im\_start$|>$assistant
\end{prompt}
\end{samepage}
\begin{samepage}
\begin{prompt}[title={Prompt Template for Self-Verification (Modified from~\citet{trust-but-verify})}]
\label{prompt: self-verification}
Below you are presented with a question and a tentative response. Your task is to evaluate the response and assign a rating to the response based on the following clear criteria:
\\
Rating Criteria:
\\
1. Missing final answer, or incorrect response with the wrong final answer: assign \verb|\|boxed\{0\}.
\\
2. Correct response with the correct final answer: assign
\verb|\|boxed\{1\}.
\\
\#\#\# Question Begin \#\#\#
\\
\{question\}
\\
\#\#\# Question End \#\#\#
\\
\#\#\# Response Begin \#\#\#
\\
\{response\}
\\
\#\#\# Response End \#\#\#
\\
First provide your evaluation process, then clearly state your final rating value enclosed in \verb|\|boxed\{\} at the end.
\end{prompt}
\end{samepage}

\end{document}